\documentclass[journal]{IEEEtran}

\usepackage[english]{babel}

\usepackage{ifpdf}

\usepackage{cite} 
\usepackage{url}
\usepackage{hyperref}
\usepackage{comment}
\usepackage{soul}

\usepackage{psfrag}
\ifCLASSINFOpdf
	\usepackage[pdftex]{graphicx}
	\graphicspath{{./figures/}}
\else
	\usepackage[dvips]{graphicx}
	\graphicspath{./figures/}
\fi
\usepackage{color}
\usepackage{xspace}
\usepackage{hyperref}
\usepackage{amsthm}
\usepackage[font=footnotesize]{caption}
\usepackage{subcaption}
\usepackage{pifont}

\newtheorem{openProblem}{Open Problem}

\usepackage[table]{xcolor} 

\newcommand{\eg}{e.g.,\xspace}
\newcommand{\ie}{i.e.,\xspace}
\newcommand{\etal}{\emph{et al.}}
\newcommand{\linkToPdf}[1]{\href{#1}{{(pdf)}}}
\newcommand{\linkToPpt}[1]{\href{#1}{{(ppt)}}}
\newcommand{\linkToCode}[1]{\href{#1}{{(code)}}}
\newcommand{\linkToWeb}[1]{\href{#1}{{(web)}}}
\newcommand{\linkToVideo}[1]{\href{#1}{{(video)}}}
\newcommand{\linkToMedia}[1]{\href{#1}{{(media)}}}
\newcommand{\award}[1]{\xspace} 

\usepackage[cmex10]{amsmath}
\usepackage{amsfonts, amssymb, amsthm}
\usepackage{mathrsfs}

\usepackage{algorithm,algpseudocode}
	\algnewcommand{\LeftComment}[1]{\Statex \(\triangleright\) #1}

\usepackage{enumerate}
\usepackage{multirow}
\usepackage{rotating}
\usepackage{booktabs}
\usepackage{lscape}

\newcommand{\edited}[1]{{\color{black} #1}}

\newtheorem{remark}{\hspace{0pt}\bf Remark}

\newtheorem{definition}{\hspace{0pt}\bf Definition}

\newcommand{\stressor}{\phi}
\newcommand{\stressors}{\phi}

\begin{document}

\title{Beyond Robustness: A Taxonomy of Approaches towards Resilient Multi-Robot Systems}

\author{Amanda~Prorok$^\star$,~
        Matthew Malencia$^\star$,~
         Luca~Carlone$^\star$,~
        Gaurav~S.~Sukhatme,~
        Brian~M.~Sadler,~
        and~Vijay~Kumar\vspace{-6mm}
        
\thanks{$^\star$These authors contributed equally. { A.\,Prorok is with the Dept. of Computer Science and Technology, University of Cambridge, email: asp45@cam.ac.uk; M. Malencia is with the University of Pennsylvania, email: malencia@seas.upenn.edu;
L.\,Carlone is with the Laboratory for Information \& Decision Systems (LIDS), Massachusetts Institute of Technology, email: lcarlone@mit.edu; G.\,Sukhatme is with the University of Southern California and Amazon (his work on this paper was performed  at  USC and is not associated with Amazon), email:  gaurav@usc.edu; B.\,Sadler is with the Army Research Laboratory, email: brian.m.sadler6.civ@mail.mil; V.\,Kumar is with the University of Pennsylvania, email: kumar@seas.upenn.edu;
}}
\thanks{This work was partially funded by ARL DCIST CRA W911NF-17-2-0181. A Prorok acknowledges the support of Physical Sciences Research Council (grant EP/S015493/1), and ERC Project 949940 (gAIa).
}
}

\maketitle

\begin{abstract}
Robustness is key to engineering, automation, and science as a whole. 
However, the property of robustness is often underpinned by costly requirements such as over-provisioning, known uncertainty and predictive models, and known adversaries. These conditions are idealistic, and often not satisfiable.
Resilience on the other hand is the capability to endure \textit{unexpected} disruptions, to recover swiftly from negative events, and bounce back to normality. 
In this survey article, we analyze how resilience is achieved in networks of agents and multi-robot systems that are able to overcome adversity by leveraging system-wide complementarity, diversity, and redundancy---often involving a reconfiguration of robotic capabilities to provide some key ability that was not present in the system a priori. 
As society increasingly depends on connected automated systems to provide key infrastructure services (e.g., logistics, transport, and precision agriculture), providing the means to achieving resilient multi-robot systems is paramount. 
By enumerating the consequences of a system that is not resilient (fragile), we argue that resilience must become a central engineering design consideration. 
Towards this goal, the community needs to gain clarity on how it is defined, measured, and maintained. 
We address these questions across foundational robotics domains, spanning perception, control, planning, and learning. 
One of our key contributions is a formal taxonomy of approaches, which also helps us discuss the defining factors and stressors for a resilient system.
Finally, this survey article gives insight as to how resilience may be achieved. Importantly, we highlight open problems that remain to be tackled in order to reap the benefits of resilient robotic systems. 
\end{abstract}

\begin{IEEEkeywords}
resilience, robustness, networked robotic systems, multi-agent systems.
\end{IEEEkeywords}

\IEEEpeerreviewmaketitle

\section{Introduction} \label{sec:intro}

\begin{figure}[tb]
    \centering
    \includegraphics[width = 0.9\columnwidth]{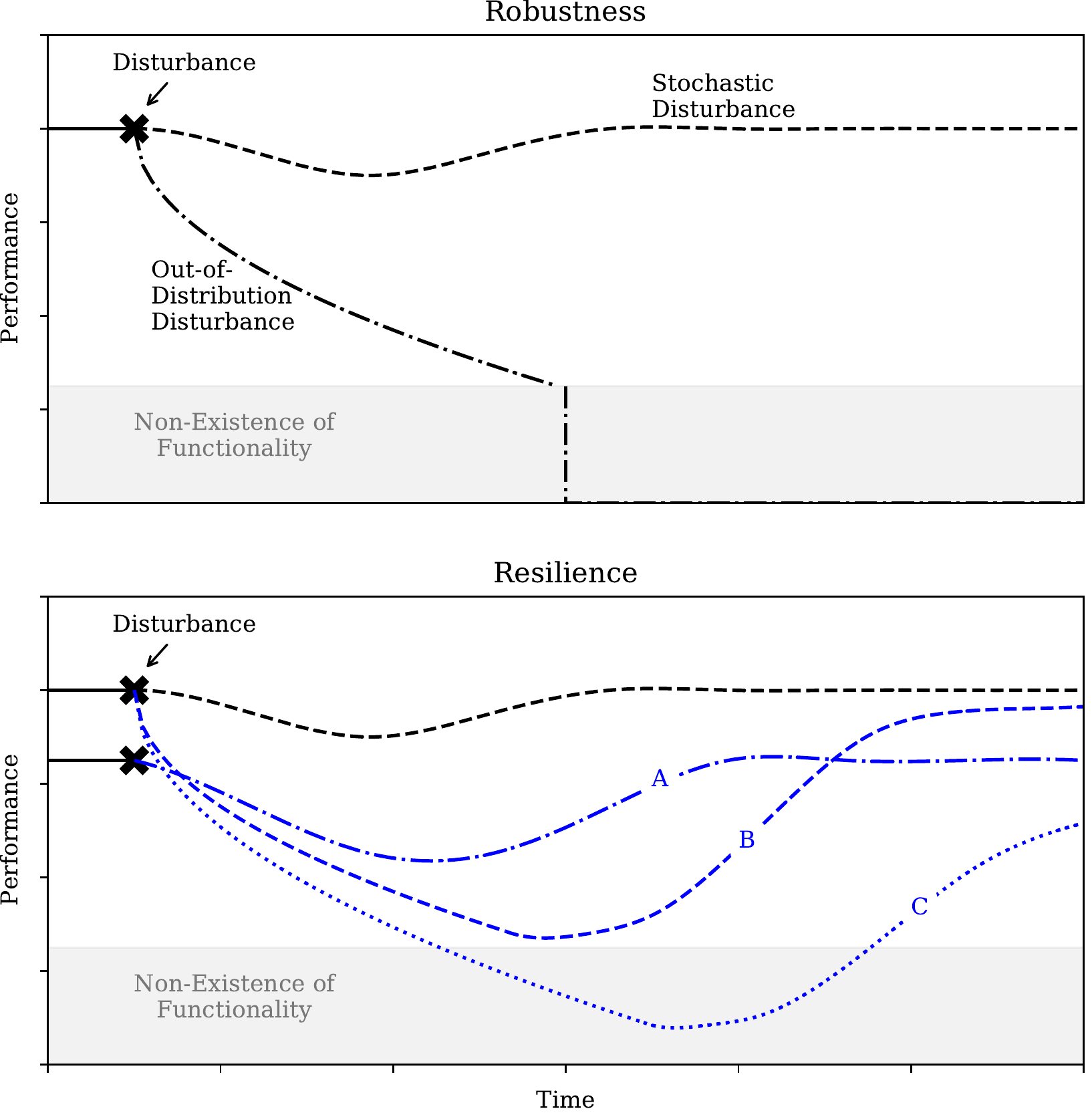}
    \caption{Robust methods (\textit{top}) aim to maintain \textit{efficiency of functionality} in the presence stochastic disturbances. However, unmodeled disruptions may lead to system-wide catastrophic failures. 
    \edited{ On the other hand, various manifestations of resilience \textit{(bottom)} can handle unexpected disturbances such as through \textit{(A)} \textit{anticipatory} policies that aim for sustained sufficiency, \textit{(B)} \textit{adaption} to recoup performance, or \textit{(C)} system-wide \textit{reorganization} to recover from loss of functionality.}
    This dichotomy is further elaborated in Section~\ref{sec:performance}.}
    \label{fig:resilience_vs_robustness}
    \vspace{-0.3cm}
\end{figure}

\textit{Robustness} is
one of the most important topics within the fields of control~\cite{sastry_Adaptive_2011}, statistics~\cite{hampel_General_1971, bradley_Robustness_1978}, engineering~\cite{baker_assessment_2008, bertsimas_Price_2004}, and, admittedly, science at large~\cite{kitano_Biological_2004, barkai_Robustness_1997}. 
Not surprisingly, robustness has also been a key objective in the design and deployment of modern multi-robot systems, from service robots for factory automation to autonomous transportation systems, where a lack of robustness at a single agent may disrupt operation for the entire system or even put human lives at risk~\cite{bradshaw_Ocado_2021}. 
\edited{However, more is needed} to incorporate ideas triggered by advances in networking technology, which has facilitated the design of connected, interdependent multi-agent systems~\cite{yang_grand_2018a}. 
\edited{Redundancy and over-provisioning have long been used to provide a degree of safety and robustness
\cite{ghare_Optimal_1969, kulturel-konak_Efficiently_2003a}; these methods, however, may lead to fragility in the face of unmodeled disruptions (see Fig.~\ref{fig:resilience_vs_robustness}, \textit{top})~\cite{bertsimas_Price_2004, doyle_robust_2005}. 
Carefully orchestrated \textit{coordination, cooperation, and collaboration} in a team of (possibly loosely) connected agents provides new opportunities~\cite{rothschild_Adam_1994}.
}

In this article, we argue that \emph{resilience} might be a property uniquely achieved through networks of agents that are able to overcome adversity by leveraging system-wide complementarity, diversity, and redundancy \edited{to \textit{preserve the existence of functionality or minimize the time-lapses in which the existence of  functionality is compromised}.
Fig.~\ref{fig:resilience_vs_robustness} \textit{(bottom)} illustrates various manifestations of resilience; such behaviors might include \textit{anticipatory} policies that aim for a course of action that is `good enough'~\cite{simon_Administrative_1947}, real-time \textit{adaption} to recoup performance loss~\cite{hassler_Resilience_2014}, or system-wide \textit{reorganization} to recover from failure~\cite{rodin_Resilience_2014a}.}
In the following, we go about defining the central notions of resilience, give insight as to how it may be achieved, and, importantly, highlight open problems that remain to be tackled in order to reap the benefits of resilient robotic systems.

\vspace{-0.3cm}
\subsection{The Need for Resilient Multi-Robot Systems}
With progress we also face new challenges. We now depend on connected automated systems to provide key infrastructural services, such as logistics~\cite{tilley_Automation_2017,kamagaew_Concept_2011}, resource distribution~\cite{enright_Optimization_2011b, ma_Lifelong_2017a}, transport systems~\cite{hyldmar_Fleet_2019b, dressler_Intervehicle_2014a, ferreira_Selforganized_2010a}, manufacturing~\cite{cherubini_Collaborative_2016}, and agriculture~\cite{noguchi_Robot_2011, albani_Monitoring_2017}.

The usage of multiple connected robots over a single robot provides evident gains (e.g., work distribution, spatial coverage, specialization). 
However, as connections are established, information is shared, and dependencies are created, these systems give rise to new vulnerabilities and threats. 
Rodin's book on resilience provides ample real-world evidence that shows how \textit{the failure of a single entity can disrupt operations to leave dependencies unanswered and fundamental  necessities unfulfilled}~\cite{rodin_Resilience_2014a}. The book argues that principles such as readiness, responsiveness, and revitalization would lead to resilience, but it is not always clear how such principles can be transformed into actionable plans. Also, while such general guidelines hold in 
{social systems, it is not clear}
how the field of automation and robotics would be able to leverage them to increase system resilience. 

{We focus on} the domain of networked robotic systems---multi-robot systems, in short---wherein individual autonomous machines work together in pursuit of higher-order missions and goals.
The virtue we seek to characterize and acquire is \textit{resilience}. Yet, how is it defined, how may it be measured? How do we build automated resilient systems? 
This survey article aims at providing answers to these questions. By doing so, our argument develops to state that {\bf{\emph{resilience must become a central engineering paradigm}}}.


\vspace{-0.3cm}
\subsection{From Robustness to Resilience}
\label{subsec:history}

\begin{figure}[tb]
    \centering
    \includegraphics[width = 0.75\columnwidth]{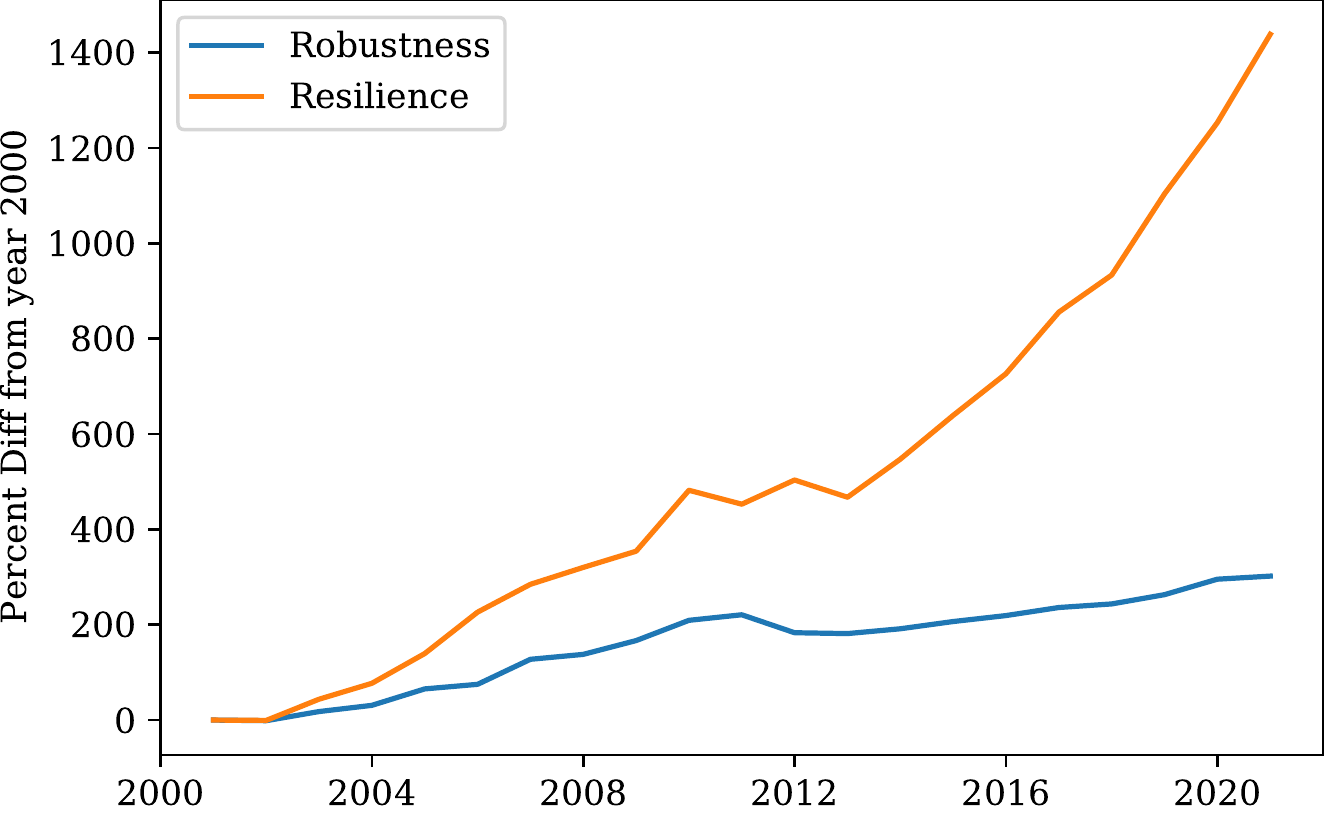}
    \caption{\edited{Publication trends reveal the rapid increase in the study of resilience, plotted as the percent difference of resilience and robustness relative to their respective number of publications in year 2000.}}
    \label{fig:resilience_pubs_xplore}
    \vspace{-0.3cm}
\end{figure}

{While robustness is a classic theme in systems engineering, resilience has emerged as an important new paradigm, and publications trends reveal this \textit{shift in focus towards resilience}.  Figure \ref{fig:resilience_pubs_xplore} plots annual publications from IEEExplore\footnote{\url{https://ieeexplore.ieee.org}}, here showing the percentage difference (relative to the year 2000) of papers with keywords `resilient' or `resilience' and of papers with keywords `robustness' or `robust'.  
Papers addressing robustness grew from just under 5,000 in 2000 to over 17,000 in 2021 (a three-fold increase), while publications considering resilience grew from 150 in 2000 to over 2,200 in 2021 (a fifteen-fold increase).}

{Robustness is central to robotics.} One of robotics' most influential papers, published by R. Brooks in 1985 (with more than 100,000 citations), is concerned with a robust robot control architecture~\cite{Brooks85}. While robustness is not explicitly tested for nor measured, given an adequate amount of over-provisioning in the robot's design, Brooks' architecture allows for real-time adjustments to  internal robot component failures.
Other early work on robustness was heavily inspired by the influence of control theorists~\cite{Slotine91book-appliedNonlinearControl}. The domain of robust robot manipulators enjoyed significant attention early on, 
\eg~\cite{slotine_Robust_1985, lim_Robust_1987, colgate_Robust_1988}, yet provided a very narrow lens on robustness through the consideration of \textit{parametric uncertainty} alone. Later work began to generalize robust measures to address causes such as imperfect motion, environmental dynamics~\cite{luders_Chance_2010}, and adversarial attacks~\cite{basilico_Extending_2009}. The commonality of these methods is their reliance on models of uncertainty and adversity, with solutions often shown to be robust under the action of \textit{bounded disturbance}. These conditions are idealistic; real-world stories abound ~\cite{taleb_black_2007}. While the field of \textit{adaptive control} aims at providing better ways of adapting to changing process dynamics through online tuning of model parameters, these methods, too, are burdened by design assumptions (e.g., gain scheduling) that restrict operations to well-defined conditions~\cite{astrom_Adaptive_2013}.

\edited{In a series of papers, Doyle and collaborators argue that any attempt to maximize robustness leads to fragility~\cite{doyle_Rules_2007, doyle_robust_2005}. This is best seen through the simple example of a linear system with a feedback loop, in which any attempt to reduce error within a range of frequencies results in an increase of error in another frequency range~\cite{olsman_Hard_2019}. 
These effects are more prominent in networked systems including cellular/molecular networks in biology~\cite{carlson_Complexity_2002} and the internet~\cite{doyle_robust_2005}.
The best designed complex networked systems are robust to random component failures, but remarkably fragile to targeted out-of-distribution attacks.}

\edited{This `robust-yet-fragile' behavior is typical of the multi-robot domain:} the failure of just one robot may cascade and consequently undermine the performance of the system as a whole (e.g., see~\cite{bradshaw_Ocado_2021}). A solution to this problem was first proposed by Parker through the ALLIANCE architecture~\cite{parker_ALLIANCE_1998} that solves multi-task problems through a distributed program that allows robots to select their actions as a function of their own internal state as well as environmental conditions. This notion of \textit{fault-tolerant} multi-robot systems was refined in a subsequent body of literature, considering specific components in the autonomy pipeline, i.e., perception, planning, and control---each of which are reviewed in depth in Sec.~\ref{sec:applications}.

The realization of diverse failure modes in the multi-robot domain instigates a delineation between robustness and resilience. We understand \textit{robustness} to be the ability to withstand or overcome adverse conditions or rigorous testing without any structural changes in the multi-robot system. Robustness accommodates uncertainty and risk, but refers to the sensitivity of a particular desirable system output in response to parametric changes or bounded disturbances. 
In contrast, \textit{resilience} refers to the capability of withstanding or overcoming adverse conditions or shocks, and unknown, unmodeled disturbances. 
Quintessentially, resilient systems \textit{relax assumptions on expected conditions}. 
Robust behaviors are mitigating, whereby actions or design choices are taken in advance of disruption (e.g., through pre-planned methods for the rejection of disturbances). 
During disruption, the system remains in the same state, as no structural transformations take place. Resilient behaviors, instead, incorporate agile policies that allow the system to transform itself, to adapt to newly perceived conditions. 
Hence, providing resilience often involves system-wide \textit{re-organization}, \textit{adaptation} and \textit{growth}.
The following definition summarizes this:
\begin{definition}[Resilient multi-robot system]
A resilient multi-robot system is capable of withstanding or overcoming {unexpected} adverse conditions or shocks, and unknown, unmodeled disturbances. The property of resilience is associated with a system-wide transformation (e.g., reconfiguration, adaptation, or growth), and refers to the contingent nature of the robots' behaviors  
\edited{that is aimed at preserving the existence of functionality or minimizing the time-lapses in which the existence of functionality is compromised.}
\end{definition}

This definition of resilience highlights the relevance of \textit{interaction} between robots so that the system as a whole can leverage emanating \textit{capabilities} with the goal of retaining some task-level \textit{performance}.
These three concepts are dealt with in more detail in Sections~\ref{sec:capabilities}, \ref{sec:interaction}, and~\ref{sec:performance}.


\vspace{-0.3cm}
\subsection{Problem Domains}
\label{subsec:domains}
In this survey, we consider three classical application domains within robotics research: perception, planning, control.
Perception is the creation of an internal model of the world given sensor data and priors.
Planning uses the robot's internal world model to plan a course of action to achieve a desired goal.
Control ensures that the course of action is correctly executed. These domains constitute three key robotic research 
areas and form the building blocks of modern autonomous systems. In this context, we use the term `perception' in a broad sense, encompassing both 2D vision (\eg object detection and pose estimation), 
3D localization and mapping, sensor fusion, and high-level scene understanding. 
Similarly, `planning' includes both motion and task planning, as well as task allocation, while 
`control' spans topics from traditional control theory to networked control and consensus.


\vspace{-0.3cm}
\subsection{Contributions of this Survey}
\label{subsec:contributions}

This article aims not only to delineate robustness vs. resilience and to contextualize this within the multi-robot domain, but also to provide actionable open problems that define directions where, we believe, we should be heading next. Our contributions are listed as follows:
\begin{itemize}
    \item We provide new \textit{definitions and terminology} that constitute the resilience problem.
    \item We provide the first \textit{formal model} of resilience engineering. We introduce notation to support the model.
    \item We provide a \textit{taxonomy} of approaches towards resilience in multi-robot systems. 
    \item We introduce key \textit{labels} that facilitate an analysis of the body of existing work and review existing papers with respect to our taxonomy and formalization. 
    \item We provide an enumeration of \textit{open problems} that encompass key challenges and areas of future work.
\end{itemize}

\edited{Following an early seminal work surveying multi-robot systems~\cite{parker_Multiple_2016},} several recent surveys on multi-robot systems have been released~\cite{Halsted21arxiv-multiRobotSurvey,Kegeleirs21frontier-multiRobotSLAMSurvey,Dorigo21ieee-multiRobotSurvey,Lajoie21arxiv-multiRobotSLAMSurvey,zhou_Multirobot_2021}, while none of them share our focus on resilience.
 Halsted, et al.,~\cite{Halsted21arxiv-multiRobotSurvey} focus on distributed optimization techniques for multi-robot systems.
 Kegeleirs, et al.,~\cite{Kegeleirs21frontier-multiRobotSLAMSurvey} focus on SLAM and point out the importance of decentralized approaches as opposed to more traditional centralized multi-robot SLAM. 
 Lajoie, et al.,~\cite{Lajoie21arxiv-multiRobotSLAMSurvey} provide a more in-depth discussion on multi-robot SLAM, including 
 mathematical formulations and discussion about open problems. 
Dorigo, et al.,~\cite{Dorigo21ieee-multiRobotSurvey} complement this survey by reviewing history and new applications of swarm robotics. 
Finally, a recent survey provided in~\cite{zhou_Multirobot_2021} has similar interests, yet focuses on a narrower segment of multi-robot approaches (i.e., mainly coordination), and does not provide formal taxonomies.


\vspace{-0.1cm}
\section{Problem Statement} \label{sec:problem}


This survey deals with the problem of designing multi-robot systems that are tasked to solve a given problem. 
Almost two decades ago, in one of the first editorials dedicated to multi-robot systems, Arai, et al., posed the following open problem~\cite{arai_Advances_2002}:
\textit{``How does the complexity of the task and of the environment affect the design of multi-robot systems?''} While posed rather broadly, this is arguably still the key question that drives this research community. Although not explicit, the question of how to design \textit{resilient} multi-robot systems is intrinsic to the original phrasing. 
We are interested in solutions to this design problem through an optimization approach which, ultimately, must be able to deal with stressors (e.g., disruptions, attacks), while striving to meet performance requirements. In the following, we elaborate the three factors that compose our design problem, i.e., \textit{(i)} robot capabilities and constraints, \textit{(ii)} the form of robot interaction, and \textit{(iii)}, performance measures.


\vspace{-0.3cm}
\subsection{Capabilities and Constraints}
\label{sec:capabilities}

In robotics, there are four fundamental \textbf{\textit{capability}} classes: \textit{(i)} sensing, \textit{(ii)} computation, \textit{(iii)} actuation, \textit{(iv)} communication~\cite{siciliano_Springer_2008}. While sensors, actuators, and computation are well-studied components that underpin an individual robot's perception-action loop, in a \textit{multi}-robot system, the action loop needs to be closed over a communication channel.
Explicit communication (e.g., via narrowband communication channels) facilitates robot interaction through the dissemination of hidden and unobservable values, giving rise to perception-action-communication loops that provide feedback to local agent controllers~\cite{yang_grand_2018a, fink_Robust_2011}.
Critical parameters include connectivity and range~\cite{mosteo_Multirobot_2008, fink_robust_2013}, topology-dependent delay~\cite{schwager_Time_2011}, and bandwidth~\cite{trawny_Cooperative_2009a,nerurkar_communicationbandwidthaware_2013}.

While tempted to design-in capabilities based on what they can do, instead, we often find ourselves limited by what they cannot do, i.e., what their {\textit{constraints}} are.
Constraints are most commonly formulated as energy budgets~\cite{robinson_efficient_2018,Tzoumas20tac-sLQG}, but specific formulations can vary: the work in~\cite{prorok_Redundant_2019a, malencia_fair_2021} includes budgets on the number of redundant robots;~\cite{setter_Energyconstrained_2016, notomista_resilient_2021} considers monitoring battery levels;~\cite{best_Online_2018} considers maximum travel time budgets;~\cite{dimario_Distributed_2015,Carlone18tro-attentionVIN} considers computing budgets; and ~\cite{liu_Optimal_2013} considers budgets that represent affordable `prices' in multi-robot auctions.
Accurately modeling constraints is reminiscent of the problem at hand, \ie resilience engineering. If constraints are characterized a priori, we can design our systems to operate accordingly. The challenge, however, lies in  discovering, adapting to, and overcoming new constraints.


\vspace{-0.3cm}
\subsection{Types of Robot Interaction}
\label{sec:interaction}

\begin{figure*}[tb]
    \centering
    \includegraphics[width = \textwidth]{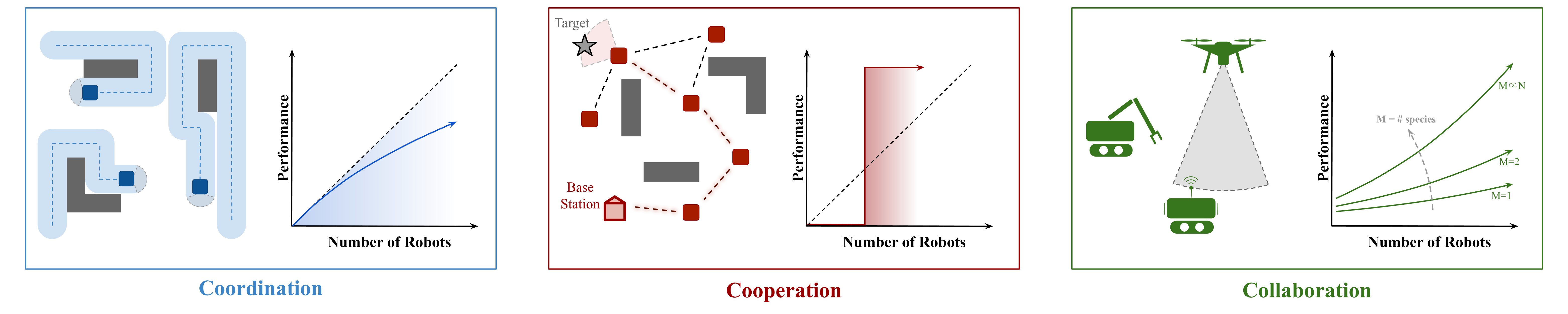}
    \caption{\edited{The \textit{three Cs} of robot interaction: coordination, cooperation, collaboration. Coordination \textit{(left)} such as multi-robot coverage, has sub-linear gains in the number of robots, where at best the gains remain proportional to the number of robots. Cooperation \textit{(middle)} achieves super-additive gains but may depend on a threshold number of robots; with a large enough team, multi-agent  search  and  tracking  problem is enabled through multi-hop communications. Collaboration \textit{(right)} involves heterogeneous agents, where performance improves with the number of species, and the corresponding complementarity of the species' capabilities.
    }}
    \label{fig:three_cs}
    \vspace{-0.3cm}
\end{figure*}

The strength of multi-robot systems lies in the robots' ability to work together. 
Networks of agents provide key infrastructural services successfully by leveraging their system-wide complementarity, diversity, and redundancy. However, not all systems interact in the same way, as dependencies arise from a variety of conditions (i.e., spatial, temporal or functional relationships). 
Different types of interaction may create different vulnerabilities or lead to different capabilities at a systems level. Here, we classify various types of multi-robot interactions into three main groups: coordination, cooperation, collaboration. We refer to these as the \textit{\underline{three Cs} of robot interaction}, illustrated in Fig.~\ref{fig:three_cs}.

\textbf{\textit{Coordination}} seeks \edited{\textit{additive} performance gains} by minimizing interference within a system, such as avoiding collisions (e.g., in multi-robot path planning) or avoiding duplicate work (e.g., in multi-robot coverage). For example, in a warehouse setting, the number of boxes that a team of robots can move per hour increases linearly as more robots join the team, as long as the team coordinates their actions. Similarly, in distributed coverage tasks, the amount of time it takes for a team of robots to cover the full area decreases linearly as more robots join the team, as long as the team coordinates their motion. 
Coordinating agents need not share goals (though they often do) because agents are awarded for their individual local performance. 
\edited{However, team performance may at times only exhibit \textit{subadditive} gains as the number of coordinating agents increases, in particular when considering cooperation among an increasingly redundant set of robots (e.g.,~\cite{prorok_Redundant_2019a}).}

\textbf{\textit{Cooperation}} considers teamwork where the system can achieve \edited{\textit{superadditive}} improvement, i.e., where the `whole is greater than the sum of its parts.’ Cooperating agents share goals and leverage teammates' help to improve task performance as a system, rather than just minimizing interference among agents as seen in coordination. \edited{Cooperation, however, may depend on threshold numbers.
Consider a multi-agent search and tracking problem in which sensed information needs to be communicated to a base-station. In addition to coordinating to enable coverage, it may be necessary for some agents to relay information using multi-hop communications. Even if more agents are recruited for the task, the performance may not increase until a communication network can be established, which may require a threshold to be exceeded. 
Thus, the performance in a team of cooperative agents may not change significantly with an increase in team size until a threshold is reached, upon which, there can be a dramatic improvement in performance. 
Similarly, in cooperative driving, a single vehicle gains no benefits on its own. As surrounding vehicles participate in a shared cooperative driving style, each vehicle gains efficiency with the help of the cooperating agents as well as gains from the increased traffic throughput that results from the reduced system-wide congestion.
On the other hand, tasks like cooperative manipulation and object transport can exhibit superadditive gains with increasing team size.}

\textbf{\textit{Collaboration}} involves \textit{heterogeneous} team interaction where agents leverage \textit{complementary} capabilities, \edited{also leading to \textit{superadditive} performance gains}. This differs from cooperation in that there is a need for specific types of agents to work together due to task requirements and inherent agent constraints~\cite{prorok_Impact_2017a}.
The resulting performance is a step function: task performance only reaches a satisfactory level when all capabilities are present.
For example, a team of agents searching for targets in a forest might leverage teamwork between aerial as well as ground vehicles. The aerial vehicles' capabilities are used to map large areas from a birds-eye view and inform exploration strategies, whereas the ground vehicles collect close-up first-person view information, or retrieve targets.


\vspace{-0.3cm}
\subsection{Performance}
\label{sec:performance}

There is abundant literature within the multi-robot field that deals with the development of methods that strive to reach and maintain efficiency (e.g., stability around an equilibrium), across a vast variety of target functionalities~\cite{siciliano_Springer_2008}.
Yet, while evidence suggests that efficiency-driven objectives lead to brittle performance under perturbation~\cite{lechner_Adversarial_2021, tsipras_Robustness_2019}, there is a dearth of work that discusses how to maintain \textit{existence} instead of \textit{efficiency} of functionality. 

Classically, robustness is measured by how much the system loses in terms of performance during disruptions. In many cases, it measures how well stability is maintained near an equilibrium state, through either the speed of return to that equilibrium or through the resistance to disturbance.
Analogously, resilience could be measured by the magnitude of disturbance that can be absorbed before the system needs to change its structure by changing the variables and processes that control behavior. But, as pointed out by Holling~\cite{holling_Engineering_1996}, there are systems that are able to maintain functionality by transitioning between multi-stable states---if there is more than one objective function, then where is the optimum, and what methods should we use to reach it?
The tension between \textit{efficiency of functionality} (e.g., thriving) and \textit{existence of functionality} (e.g., surviving) is still poorly understood, and few measures exist to quantify it. The dichotomy between robustness and resilience is illustrated in Fig.~\ref{fig:resilience_vs_robustness}.

A few recent works propose domain-specific measures of resilience. For example, in transport engineering, resilience is estimated as the change in efficiency resulting from roadway disruptions~\cite{ganin_Resilience_2017}. Areas such as biology~\cite{pimm_complexity_1984}, health~\cite{zautra_Resilience_2010, davydov_Resilience_2010}, and the built environment~\cite{hassler_Resilience_2014} have also dedicated a decade of research into this broad question; but tying together formalisms in a cross-disciplinary manner proves hard, if not impossible, due to incompatible quantities of interest. While preceding ideas may, at the very least, inspire resilience measures in the multi-robot systems domain, further dimensions must be considered: \eg the various behavioral changes that occur over time to stymie a disruption, or the time it takes for the system to reach a new steady state, or the performance of the system at the equilibrium after disruption, or even the number of possible equilibria that ensure system functionality.

\begin{openProblem} [Measurement of Resilience]
If the property of resilience is associated with a system-wide transformation (e.g.,  reconfiguration,  adaptation,  or  growth), then new multi-dimensional measures need to be developed that account for these changes holistically. 
\label{op:resilience_measure}
\end{openProblem}


\vspace{-0.3cm}
\section{Taxonomy of Approaches} \label{sec:approaches}


We posit that resilience in multi-robot systems is achieved through the joint execution of a strategy that aims to overcome the adverse effects of undesirable disturbances. Because there exist multiple types of disturbances (i.e., \textit{stressors}) and multiple means to addressing each stressor, work in multi-robot resilience can vary greatly. This section outlines the \textit{key variables} defining each \textit{stressor} type, the \textit{approach} types for tuning the key variables, and the relationship between {stressors} and {approaches}.


\vspace{-0.3cm}
\subsection{Stressors and Key Variables}

Networked robotic systems encounter myriad adverse conditions, such as distributional noise whose instantiation is a priori unknown~\cite{lajoie_DOORSLAM_2020,Mangelson18icra}, and disturbances outside the robots' world model~\cite{lechner_Adversarial_2021}. We also consider targeted disturbances, such as adversaries intent on disrupting the system (e.g.,~\cite{saulnier_Resilient_2017a, mitchell2020gaussian}) and non-cooperative agents competing for resources~\cite{lowe_MultiAgent_2017a, blumenkamp_Emergence_2020b}. Resilience is achieved by withstanding and overcoming such adverse conditions. We identify two main \textit{stressor} dimensions that delineate whether the stressor is stochastic or out-of-distribution, and whether the stressor is targeted or not. Each stressor type is defined by \textit{key variables,} i.e., the models and parameters that characterize it. 

\textbf{ \textit{Stochastic}} stressors entail the noise and uncertainty that is present throughout robotics systems. The key variables of stochastic stressors are hyper-parameters of a disturbance model. For example, the key variables of a Gaussian stochastic stressor are mean and standard deviation. When dealing with stochastic stressors, the system may possess a priori knowledge of the type of model that characterizes the disturbance, but may also be capable of adapting or updating the model parameters in-situ, during operation. Even with a perfect model and well-tuned key variables, robotic systems are challenged because the exact, true instantiation of such stressors is rarely known a priori. 

\textbf{ \textit{Out-of-distribution}} stressors are disturbances that are not captured by the robot systems' model. Similar to stochastic stressors, the exact, true instantiation of an out-of-distribution stressor is rarely known a priori. However, out-of-distribution stressors are more challenging because the disturbance is not even probabilistically known beforehand; in other words, the disturbance is unknown to the model. Therefore, the key variable of out-of-distribution stressors is the model itself, together with its hyper-parameters (which may change when the model changes). 

\textbf{ \textit{Targeted}} stressors are distinguished by the existence of an \textit{intent}: targeted stressors are goal-oriented and are therefore often functions of the robotic system itself, capable of adapting to changes in the robotic system. A common example of targeted stressors are adversarial disturbances, which are intent on disrupting the robotic system. But not all targeted stressors are adversarial---external agents (that have their own goals and are therefore non-cooperative) compete for the resources that are shared with the robotic system in question. For example, connected autonomous multi-vehicle systems that share road-space with human drivers must deal with their potentially non-cooperative driving behavior. The humans' behavior is targeted (i.e., through egocentric driving goals), yet non-adversarial.

\begin{figure}
    \centering
    \includegraphics[scale=0.35]{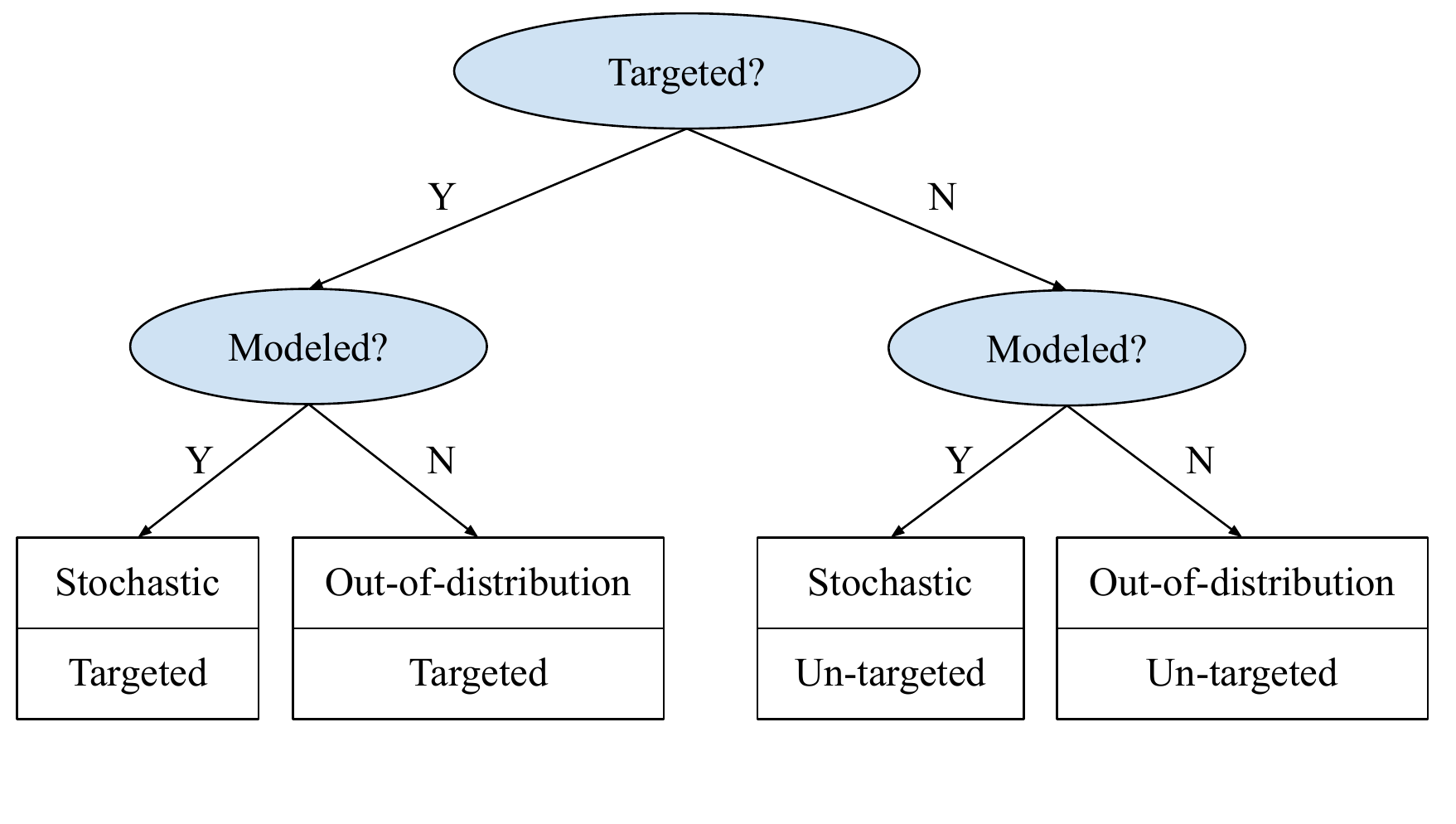}
    \caption{The classification of stressor types, showing that the stressors can be either targeted or un-targeted. Further, the stressors can be classified as stochastic, if the robots have a model of the stressor, or out-of-distribution otherwise.}
    \label{fig:stressors}
    \vspace{-0.3cm}
\end{figure}

The distinction of targeted and untargeted stressors does not impact the key variables. As seen in Fig.~\ref{fig:stressors}, targeted and untargeted stressors are both further categorized as stochastic versus out-of-distribution stressors which in turn define the key variables (as seen above).

\vspace{-0.3cm}
\subsection{Approaches}

There are multiple types of approaches by which a robotic system can withstand or overcome stressors. Stressor types, as described above, are defined by their key variables. Similarly, approach types are defined by how robotic systems interact with key stressors variables. Specifically, a stressor type defines \textit{what} the key variables are, whereas an approach type defines \textit{how} the key variables are tuned. In the following, we introduce three approach types, \textit{\textbf{pre-}}, \textit{\textbf{intra-}}, and \textit{\textbf{post-operative}}, which are agnostic to the stressor type.

To help define these approaches, we make use of some notation. In robotics applications ranging from estimation to planning, the goal is to optimize an objective function\footnote{We illustrate this optimization with the \textit{max} function, though any optimization function can be used.} (e.g., localization accuracy or coverage) over some time horizon. Let $f$ represent the \textbf{objective function}, \edited{which takes a general form here to encapsulate the various manifestations of resilience seen in Fig.\ref{fig:resilience_vs_robustness} and the problem domains considered in this survey}. Let $x$ represent the system decision variable over which this function is optimized, and $T$ be the time horizon (in non-sequential optimization $T$ is unit step). Resilience in these systems is concerned with the impact of the key variables of the stressors, denoted $\stressors$, on the system performance, i.e., the existence of functionality.

\textbf{ \textit{Pre-operative}} approaches pre-determine the key variables of the stressor \edited{a priori, and provide resilience \textit{by design}}. In the context of the robotics system, this means the objective function is optimized with respect to the decision variable assuming given (or pre-calculated) key variables.

\begin{definition}[Pre-operative] \label{def:pre-operative}
Pre-operative approaches optimize the system objective function, $f$, over the system decision variables, $x$, with respect to given key variable values, $\stressors$:

\[
\displaystyle{ \max_{x_0 ... x_T} \ f(x \ | \ \stressors).}
\]
\end{definition}

Because these approaches are offline with respect to the stressor, they often provide resilience against an expected or worst-case disturbance~\footnote{This definition of pre-operative approaches is reminiscent of \textit{robustness}}. 
One trait of pre-operative approaches is that their resilient actions are taken regardless of the presence of the stressor. For example, a system that is designed to be resilient to five adversaries will take the same actions no matter how many adversaries are present. Another common aim of pre-operative approaches is to achieve robustness or resilience through over-provisioning and redundancy.

\textbf{ \textit{Intra-operative}} approaches are online with respect to the key variables of the stressor, \edited{and provide resilience \textit{through adaptation}}. During operation, changes to the key variables are made, therefore, the stressor key variables $\stressors$ become decision variables in the optimization.

\begin{definition}[Intra-operative] Intra-operative approaches optimize the system objective function, $f$, over \textit{both} the system decision variables, $x$, and stressor key variable values, $\stressors$, such that the stressor is addressed online:
\[
\displaystyle{\max_{\stressors, \ x_i ... x_{i+T}} \ f(x, \stressors) }.
\]
\end{definition}
Intra-operative approaches often use online algorithms with respect to the decision variable $x$ (i.e., both the system decision variable and the stressor key variables are updated online), but note that the opposite is not necessarily true. Some algorithms may be online with respect to the system variable, $x$, but assume a fixed key variable, $\stressors$, throughout; therefore, they are pre-operative because they are offline with respect to the stressor. In contrast to pre-operative approaches, intra-operative approaches will not take resilient actions when there is no stressor present.

\textbf{ \textit{Post-operative}} approaches update the key variables of the stressor using past data (often in batch), \edited{and provide resilience \textit{through learning}}. Therefore, the system variable $x$ is no longer a decision variable. Rather, the optimization seeks the best key variables given a set of data. 

\begin{definition}[Post-operative] \label{def:post-operative} Post-operative approaches optimize the stressor key variables, $\stressors$, given past data of the system objective function, $f$, and system decision variables, $x$:
\[
\displaystyle{\mathrm{argmax}_{\stressors} \ f(\stressors \ | \ x_{i-T} ... x_i).}
\]
\end{definition}
Similar to pre-operative approaches, post-operative approaches are offline with respect to the stressor, with the difference that tuning is executed \textit{post-factum}. In other words, post-operative approaches seek improvement in future trials, \textit{after} the robot system's performance on relevant tasks has been experienced and measured. The unique feature of post-operative approaches is that they facilitate the discovery of out-of-distribution stressors. Examples of post-operative approaches include co-design, evolutionary optimization, and learning techniques (e.g., off-policy, lifelong, reinforcement learning).

\edited{

\begin{remark}[Objective function notation]
The general notation used for the objective functions $f$ in Definitions~\ref{def:pre-operative}-\ref{def:post-operative} can represent the many manifestations of resilience seen in Fig.~\ref{fig:resilience_vs_robustness}. Current work, and many of the citations in Table \ref{taxonomy}, often interpret these objective functions in the form of optimization objectives where resilience is scalarized, as is the case with measures of efficiency. However, the general notation of $f$ encapsulates a broad class of functions, with the ability to incorporate constraints; model hybrid systems and indicator functions; capture system survival, such as defining a survival threshold $f_0$ and an objective function of the form $f(x, \stressor) \ge f_0$; and represent the measures of resilience that may result by future research addressing Open Problem \ref{op:resilience_measure}.  
\end{remark}}

\vspace{-0.3cm}
\subsection{Stressor and Approach Relationship}

The relationship between stressor types and approach types is straightforward: the approach type defines how the key variables are updated while the stressor type defines the key variables of interest. To add clarity and comprehensiveness, we build on the above notation used to define the approach types and introduce new notation for the stressors. While $\stressors$ abstractly represents any stressor, we introduce more specific notation in order to distinguish between the different stressor types. Let $\theta$ be the hyper-parameters of the stressor and $g$ be the model of the stressor, both of which apply to targeted and untargeted stressors.  

Fig.~\ref{fig:key_var} shows that stochastic stressors are defined and tuned by their hyper-parameters $\theta$ while out-of-distribution stressors are governed by their model $g$ as well as the hyper-parameters. Furthermore, out-of-distribution stressors cannot be addressed by pre-operative approaches. 
To be pre-operative, the stressor model would need to encompass the disturbance; if the model does contain the disturbance, then the stressor is stochastic rather than out-of-distribution, whereas if the model does not contain the disturbance, then no resilience has been achieved and the disturbance must be handled in an intra- or post-operative manner.

\begin{figure}[]
    \centering
    \begin{tabular}{c|cc}
    \hline  
    Approach & Stochastic & Out of Dist \\
    \hline 
    Pre-operative & Offline $\theta$ & N/A \\
    Intra-operative & Online $\theta$ & Online $g,\theta$\\
    Post-operative & Update $\theta$ & Update $g, \theta$ \\
    \hline
    \end{tabular}
    \caption{The relationship between approach types and stressor types, showing how each approach changes or uses the key variables for each stressor type.}
    \label{fig:key_var}
    \vspace{-0.3cm}
\end{figure}


\vspace{-0.3cm}
\section{Application Domains} \label{sec:applications}


This section tailors the notion of resilience to three key application domains in robotics: 
perception (Section~\ref{subsec:perception}), 
planning (Section~\ref{subsec:planning}),  
and control (Section~\ref{subsec:control}).  
In each subsection, 
\textit{(i)} we discuss typical stressors and multi-robot interactions, 
\textit{(ii)} classify existing approaches into pre-operative, intra-operative, and post-operative, 
and \textit{(iii)} highlight open problems. 
We conclude the section with a short review of other domains 
(Section~\ref{subsec:others}), including robot co-design.


\vspace{-0.3cm}
\subsection{Perception and Estimation}
\label{subsec:perception}

Perception ---the robot’s ability to sense and understand the surrounding environment--- is a key enabler for autonomous systems’ operation in complex environments, and provides functionalities such as estimating the location of the robot, building a map of obstacles in its surroundings, detecting, classifying, and tracking objects. 
This capability is even more crucial for multi-robot systems, where a shared understanding of the world is 
 a key requirement for successful 
interaction. 

However, multi-robot systems pose new challenges to perception: 
(i) the sensor data is collected independently by multiple robots, possibly equipped with different sensor suites and with limited onboard compute, 
(ii) the team needs to form a shared world model in the face of communication constraints (\eg bandwidth, communication range, privacy), and (iii) the scale of multi-robot perception problems exacerbates 
the limitations that already arise in single-robot perception (\eg scalability, noisy and out-of-distribution measurements).

In the following, we review perception problems arising in multi-robot systems, spanning several  
subdomains (\eg 
low-level perception and 2D vision,
localization and mapping,
and high-level scene understanding). 
Note that we restrict our focus to \emph{spatial} perception (\ie we are mainly concerned with estimating 
quantities that live in 3D space), and do not cover other perception problems (\eg action and emotion recognition) nor 
prediction problems.

\textbf{Pre-operative approaches.} We review pre-operative approaches for 
(i) low-level perception, (ii) localization, mapping, and estimation, and (iii) describe open problems in high-level
multi-robot learning and real-time scene understanding.

\subsubsection{Low-level Perception and 2D Vision}
Low-level perception focuses on image --or more generally sensor's signal-- processing and is typically finalized to 
detecting features or objects, performing pixel-wise semantic segmentation, 
and recognizing known places, among other problems. Low-level perception methods are often referred to as the \emph{perception front-end}~\cite{Cadena16tro-SLAMsurvey}. In these problems, common stressors include
illumination and viewpoint changes, presence of unexpected dynamic elements in the scene, and in general the presence of 
nuisances that are irrelevant for the perception task. 
The following approaches are classified as pre-operative since they do not adapt during operation, 
are often designed for the worst case, or do not explicitly deal with stressors.

Multi-robot research has extensively investigated \emph{distributed place recognition}, where 
robots in a team have to detect whether they are observing the same place; place recognition 
enables re-localization, and loop closure detection in Simultaneous Localization and Mapping (SLAM)~\cite{Cadena16tro-SLAMsurvey}. 
In a centralized setup, a common way to obtain loop closures is to use visual
place recognition methods, which compare compact image descriptors to find
potential loop closures. This is traditionally done with global visual
features~\cite{Arandjelovic16cvpr-netvlad}, or local visual
features which can be quantized in a bag-of-word
model~\cite{Sivic03iccv}. The feature descriptors are designed to gain robustness 
to the stressors (\eg viewpoint changes). 
Distributed loop closure detection aims at detecting loop closures without exchanging raw data, 
a desirable feature when the robots operate under range and bandwidth constraints.
Tardioli, et al.,~\cite{Tardioli15iros} use visual vocabulary indexes instead of descriptors to reduce the required bandwidth.
Cieslewski and Scaramuzza~\cite{Cieslewski18icra} propose distributed
and scalable solutions for place recognition in a fully connected team of
robots, using bag-of-words of visual features~\cite{Cieslewski17ral-bow} or 
full-image NETVLAD descriptors~\cite{Cieslewski17mrs-netvlad}. 
Tian, et al.,~\cite{Tian18rss}, Lajoie, et al.,~\cite{Lajoie20ral-doorSLAM}, and
Giamou, et al.,~\cite{Giamou18icra} propose approaches to coordinate the
data exchange during the geometric verification step.

Recent effort in computer vision has focused on \emph{segmentation and recognition problems}.
Liu, et al.,~\cite{Liu20cvpr-when2com} learn to construct
communication groups and decide when to communicate for
multi-agent semantic segmentation and 3D shape recognition tasks. 
Wu, et al.,~\cite{Wu19iccv-RLVideoRecognition} use multi-agent reinforcement learning to 
sample frames that maximize the accuracy of video recognition.
Mousavi, et al.,~\cite{Mousavi19iros-multiAgentImageClassification} propose 
a  multi-agent image classification approach based on generalized policy gradient.

Low-level perception problems are often performed using learning-based techniques, including 
descriptor learning for place recognition~\cite{Arandjelovic16cvpr-netvlad}.
While currently less used in robotics, the growing field of \emph{federated learning} investigates how to train machine learning models in a distributed fashion, with the goal of preserving privacy of the agents 
in the team and sharing computational resources~\cite{Konecny18iclr-federatedML}.
This field is still in its infancy, with recent effort being devoted to dealing with unreliable agents or unreliable connectivity~\cite{Salehi21toc-federatedML}.  
While most approaches listed in this section do not formally address the presence of stressors, the 
literature on \emph{adversarial learning} attempts to quantify and improve the robustness of a neural network to {perturbations}, see~\eg~\cite{Madry17arxiv-adversarialML}.
Most of these approaches consider simple perturbations (\eg additive pixel-wise noise on an image) that lead the network \mbox{to produce incorrect classifications.}

\subsubsection{Localization, Mapping, and Estimation} 
Here we briefly review distributed estimation techniques and then focus on their applications in multi-robot teams. 
Estimation techniques typically constitute the \emph{perception back-end}~\cite{Cadena16tro-SLAMsurvey}, in that 
they take intermediate representations produced by low-level perception processes (the perception front-end) 
and use them to estimate the state of the system (\eg the pose of the robots, the 3D location and velocity of 
objects in the environment). 
In these problems, typical stressors include measurement noise, out-of-distribution data (typically produced by incorrect processing at the front-end or by off-nominal sensor behavior), as well as intermittent communication. 

Early work focuses on estimation with Gaussian noise, a setup that builds on well-established estimation-theoretic methods~\cite{Mendel95book,Dellaert17fnt-factorGraph}. 
Distributed estimation in multi-agent systems has also been extensively investigated in robotics and 
sensor networks, with the goal of developing methods that converge to optimal estimates while only requiring local 
communication~\cite{Barooah07csm} 
and are possibly robust to unreliable communication channels~\cite{Schenato07ieee}. 
Multi-robot research 
investigates multi-robot localization with different estimation
techniques, including Extended Kalman filters~\cite{Roumeliotis02tra}, 
information filters~\cite{Thrun03isrr}, and particle filters~\cite{Howard06ieee,Carlone10jirs-multiRobotSLAM}.
Maximum a posteriori and maximum likelihood estimation have recently been
adopted as a general and accurate framework for robotics; in SLAM problems, 
these frameworks lead to well-studied optimization problems, including \emph{pose graph optimization} (PGO)~\cite{Cadena16tro-SLAMsurvey} or \emph{factor graph optimization}~\cite{Dellaert17fnt-factorGraph}.
Early literature on multi-robot PGO focused on centralized approaches, where measurements 
are collected at a central station, which computes the trajectory estimates for all the
robots~\cite{Andersson08icra,Kim10icra,Bailey11icra,Lazaro11icra,Dong15icra}.
{Since the computation workload and the communication bandwidth of a
centralized approach grow with the number of robots, related work has 
explored \emph{distributed techniques}, in which robots perform local communication and share
the computational workload~\cite{Aragues11icra-distributedLocalization,Cunningham13icra,Choudhary17ijrr-distributedPGO3D,tian_Asynchronous_2020}; these techniques leverage problem structure and distributed optimization methods
to obtain optimal estimates from partial information exchange. 
 The works~\cite{Forster13iros-airGroundLocalization,Michael14fr-airGroundMapping} consider a collaborative setup with ground and aerial robots.

Recent work on multi-robot localization, mapping, and estimation has focused on 
the realistic case where some of the measurements used by the back-end are \emph{outliers}
(\ie they are affected by severe unmodeled noise). 
The fundamental problem of robust estimation has a long history and there are 
well established frameworks to model estimation problems with outliers, including 
M-estimation and consensus 
maximization~\cite{Chin18ECCV-robustFitting,Antonante21tro-outlierRobustEstimation}. 
However, these frameworks typically lead to hard optimization 
problems~\cite{Chin18ECCV-robustFitting,Antonante21tro-outlierRobustEstimation}, and developing 
fast and effective solvers is still an active research area~\cite{Yang20neurips-certifiablePerception,Yang20ral-GNC,Barron19cvpr-generalAdaptiveLoss,Chebrolu2020arxiv-adaptiveRobustKernels}.
We remark that these approaches are still pre-operative (according to Definition~\ref{def:pre-operative}): for instance
M-estimators can be understood as maximum-likelihood estimators over heavy-tailed (but known) noise models.
Robust estimation is particularly important in multi-robot localization and mapping where 
incorrect measurements among the robots are more difficult to detect when the robots do not share a common reference frame.
Centralized outlier rejection techniques for multi-robot SLAM include 
voting schemes~\cite{Indelman14icra} and graph-theoretic methods~\cite{Mangelson18icra}; 
the Pairwise Consistency Maximization approach of~\cite{Mangelson18icra} has been particularly successful, 
with a field deployment reported in~\cite{Ebadi20icra-LAMP}, and a distributed implementation proposed in~\cite{Lajoie20ral-doorSLAM}. 
More recently, Tian, et al.,~\cite{Tian21arxiv-KimeraMulti} propose a distributed M-estimation approach based on graduated non-convexity~\cite{Yang20ral-GNC} which is shown to lead to more accurate trajectory and map estimates.

\begin{openProblem}[Learning in teams] \label{op:team_learning}
Federated and adversarial learning have the potential to enhance multi-robot operation 
but have found limited use in robotics.
Open challenges in federated learning for multi-robot systems include improving
communication bandwidth, energy efficiency, and security~\cite{Aledhari20access-federatedLearning}. 
Regarding adversarial learning, robotics and computer vision applications 
require going beyond simple additive perturbation models, which are 
not well-suited to capture nuisances arising in real perception problems~\cite{Poursaeed18cvpr-adversarialML}.
\end{openProblem}

\begin{openProblem} [Distributed real-time scene understanding]
\label{op:scene_understanding}
While multi-robot SLAM can be considered a mature field of research, the goal of achieving human-level 
understanding of the environment is still out of reach for a robot. 
Despite the growing literature on single-robot metric-semantic understanding (\eg~\cite{McCormac17icra-semanticFusion,Bowman17icra}) and 
3D scene graph representations~\cite{Rosinol20rss-dynamicSceneGraphs,Armeni19iccv-3DsceneGraphs}, 
few papers have considered metric-semantic multi-robot mapping~\cite{Tian21arxiv-KimeraMulti,Tchuiev20ral-semanticMultiRobotMapping,Yue20iros-semanticMultiRobotMapping}. 
Infusing semantic and high-level understanding in localization and mapping problems creates novel opportunities to 
improve resilience since a team of robots can dynamically adjust depending on the external context or the semantic 
elements in the scene (\eg presence of a threat). Moreover, it  allows creating a distributed \emph{spatial knowledge base}, 
which can support several tasks from human-robot interaction to long-term autonomy. 
\end{openProblem}

\textbf{Intra-operative approaches.} 
The literature on intra-operative approaches to perception is more sparse but growing.

\subsubsection{Low-level Perception and 2D Vision}
Learning-based approaches for low-level perception (\eg object detection) are challenged by 
(i) a potential shift between the training and testing distributions, and (ii) test instances 
that belong to the tails of the distribution (\eg rare examples) for which little training data is available.
Intra-operative approaches include methods that deal with these challenges online during operation.  
The survey by Abass, et al.,~\cite{Abbass20vc-onlineLearningTracking} provides an extensive review of online learning for visual tracking. 

Recent work in robotics and autonomous vehicles uses learning-based methods to detect out-of-distribution 
examples online during operation.
 Rahman, et al.,~\cite{RahmanIROS19-falseNegative} process the hidden layer outputs of a neural 
 network to predict when a traffic sign detection network outputs false negatives. 
Henzinger, et al.,~\cite{Henzinger20ecai} observe neuron activation patterns to monitor when the network is operating on inputs unlike the data seen during training.
  Hendrycks, et al.,~\cite{Hendrycks16iclr} develop a method of monitoring network confidence based on softmax probabilities. Gupta and Carlone~\cite{Gupta20itsc-atom} propose
  Adversarially-Trained Online Monitor (ATOM) to flag incorrect detections of pedestrians in self-driving applications.
System-level monitoring approach are studied in~\cite{Antonante21iros-perSysMonitoring2} to detect off-nominal behaviors of perception modules.

\subsubsection{Localization, Mapping, and Estimation} 
We start by remarking that several approaches for robust estimation admit an alternative interpretation 
as intra-operative approaches. For instance, approaches based on graduated non-convexity, reweighted least squares, 
and dynamic covariance scaling~\cite{Yang20ral-GNC,Antonante21tro-outlierRobustEstimation,Sunderhauf13icra,Agarwal13icra} 
can be understood as approaches to adjust the measurement covariances (a key stressor variable, see Table~\ref{fig:key_var}) online, to down-weight out-of-distribution measurements. The connection between robust estimation and measurement weighting is a well-understood one and goes back to the seminal work from Black and Rangarajan~\cite{Black96ijcv-unification}, with more recent multi-robot applications in~\cite{Tian21arxiv-KimeraMulti}.

More recent work on robust estimation for robust localization, mapping, and learning goes further and 
explicitly tackles online adaptation. Antonante, et al.,~\cite{Antonante21tro-outlierRobustEstimation} propose minimally tuned robust estimation algorithms that can learn the inlier noise statistics online. 
Barron~\cite{Barron19cvpr-generalAdaptiveLoss} and Chebrolu, et al.,~\cite{Chebrolu2020arxiv-adaptiveRobustKernels} 
adjust the choice of robust loss function (within a parametric family) using an automatic online tuning procedure.

Other potential stressors include sensor mis-calibrations and sensor failures.
 Online calibration has been extensively investigated in the context of 
 kinematic odometry~\cite{Roy99icra-onlineCalibration}, 
  visual-inertial~\cite{Lee20iros-onlineCalibration} and lidar-inertial odometry~\cite{Tagliabue20iser-lion}, 
 and SLAM~\cite{Nobre17iser-selfCalibration}.
 System-integration efforts have also investigated system reconfiguration in response to sensor 
 failures~\cite{Palieri21ral-locus}. A general framework for sensor selection based on resilient submodular maximization is investigated in~\cite{Tzoumas18cdc}.

\begin{openProblem}[Resilience and reasoning over failures]
\label{op:failure_reasoning}
At the algorithmic level, resilient perception is still in its infancy: most perception frameworks 
are ``rigid'' and target robustness rather than resilience and online adaptability. 
 It is desirable for future perception algorithms to perform automatic parameter tuning to adjust to 
 heterogeneous environmental conditions.
At the system level, monitoring of perception systems is also a largely unexplored topic: how to detect failures 
of perception algorithms? how to detect that the world models built by different robots in a team 
are inconsistent with each other? 
More importantly, robot perception currently aims at \emph{detecting and isolating} off-nominal data (\eg outliers, sensor failures, 
algorithmic failures) rather than reasoning on the cause of those failures and learning how to avoid them in the 
future.
\end{openProblem}

\begin{openProblem} [Task-dependent perception and active perception]
\label{op:active_perception}
As already stressed in~\cite{Cadena16tro-SLAMsurvey}, an open-challenge is to develop a tractable
and general framework for \emph{task-driven perception}, which can guide sensing and perception processes to 
maximize a task-driven performance metric (\eg obstacle avoidance) while minimizing computation, sensing, or communication. This is particularly important in multi-robot teams, where --under communication constraints-- it is desired for the robots to exchange the minimum amount of information to guarantee successful completion of a task.
 Conversely, resilience also requires active perception, \ie how to actively plan and control the robot to 
 minimize the impact of environmental stressors.
\end{openProblem}

\textbf{Post-operative approaches.}
Postoperative approaches use batch training data collected by a robot over multiple deployments to identify stressors. 
Approaches for offline system identification and sensor calibration fall in this category, 
see, \eg~\cite{Furgale13iros}; in this case, the training is often augmented with external sensors (\eg a vicon system) to increase the observability of the resulting parameter estimation problem. 

More recently, post-operative approaches have focused on learning-based methods that can improve and adapt after multiple 
executions. In this sense, post-operative approaches are related to \emph{domain adaptation and transfer learning} in machine learning, where the goal is to allow a network --trained on a given training distribution-- to transfer to a different test distribution. 
Domain adaptation can rely on external supervision, but can also be semi-supervised or unsupervised.

For future operation of multi-robot teams, the unsupervised (or self-supervised) setup is particularly appealing since 
it avoids massive human annotations~\cite{Jing20pami-selfsupervised}. 
Self-supervision has been proven useful to learn depth, 
 optical flow, 
 visual odometry, 
 and feature descriptors for scan matching. 

\begin{openProblem}[Tuning and reconfiguration]
\label{op:tuning_reconfig}
While offline calibration is well understood, currently there are no efficient and automatic ways  
to automatically tune parameters and potentially reconfigure components in complex perception pipelines.
For instance, modern SLAM and VIO pipelines include tens to hundreds of tunable configuration parameters (\eg number of features, 
type of feature descriptors, etc.) that impact performance, 
are scenario-dependent, and rely on manual tuning from an expert. 
The large number of parameters (and the potential lack of ground-truth information) quickly makes brute-force and black-box 
approaches for tuning (\eg Bayesian optimization) impractical. 
This adds to the combinatorial complexity of choosing how to combine different algorithmic blocks comprising the robot 
perception system: which object detector? which 3D object pose estimation and tracking approach? which SLAM pipeline?   
\end{openProblem}




\vspace{-0.3cm}
\subsection{Planning and Task Assignment}
\label{subsec:planning}
Planning and task assignment are fundamental problems in multi-robot systems. Teams of robots must collectively optimize the assignment of mobile robots to tasks~\cite{kuhn_hungarian_1955}, plan schedules and action sequences that are conflict-free~\cite{torreno_cooperative_2017}, and route individual agents along collision-free paths \cite{atzmon_probabilistic_2020}. These planning problems arise in many applications, including
product pickup and delivery~\cite{grippa_Drone_2019a, jorgensen_team_2017}, item retrieval in warehouses~\cite{enright_Optimization_2011b, peltzer_stt-cbs_2020}, and mobility-on-demand services~\cite{alonso-mora_Ondemand_2017a, salzman_research_2020}.
 
Planning entails optimizing higher level goals, such as minimizing the cost of an assignment \cite{nam_analyzing_2017} or the average travel time among agents \cite{prorok_redundant_2019}. To orchestrate this coordination, centralized communication architectures have become the norm in various instances; a centralized unit collects all costs (e.g., expected travel times) to determine the optimal plan or assignment (e.g., through search algorithms such as RRT or the Hungarian algorithm). However, \emph{the optimality of this assignment hinges on the accuracy of the assignment cost estimates}. 

Despite best efforts to model any uncertainties, discrepancies between model assumptions and real-life dynamics may arise \cite{nam_when_2015}. For example, in transport scenarios, a robot may encounter an unexpectedly blocked path, and consequently takes significantly longer to reach its destination than anticipated \cite{prorok_Redundant_2019a}. Travel time uncertainty also arises due to the deterioration of positioning accuracy (e.g., GNSS service deterioration). Furthermore, recent methods consider it \emph{desirable} to actively obfuscate true robot state information (e.g., robot positioning), to ensure privacy across a variety of applications~\cite{prorok_Privacypreserving_2017a}.
Irrespective of the source of uncertainty, it follows that any discrepancies around true robot states cause a degradation in the system's overall performance, and can lead to cascading effects. Furthermore, multi-robot systems have uncertainties beyond individual robot states. The strengths of multi-robot collaboration are juxtaposed with added disruptions. To achieve resilient performance, networked robotic systems must not only cope with a higher likelihood of individual robots among a large team failing, but also with compounding uncertainties among team members, second order effects, and the impact of real world complexities on collective planning.

\textbf{Pre-operative approaches.}
Multi-agent planning can generally be categorized into assignment \cite{yang_algorithm_2020}, routing \cite{toth_vehicle_2002}, or path planning \cite{stern_multi-agent_2019}. In each of these categories, pre-operative approaches create team plans offline, making decisions about the whole team's actions a priori. Although these pre-operative approaches create plans before the existence of any disturbance, they can still plan for disturbances that they modeled. Take for example the multi-agent path planning problem. Following the notation in Section~\ref{sec:approaches}, the planning algorithm seeks to optimize the average travel time $f$ given a model of uncertainty due to traffic $\stressors$ by searching over the space of paths $x$.

\subsubsection{Assignment}
Assignments under random costs have gained a considerable amount of attention~\cite{nam_When_2015a, nam_Analyzing_2017a, ponda_distributed_2012, shang_stochastic_2020}. The focus has primarily been on providing analyses of the performance under noisy conditions. Prorok and Kumar consider privacy in mobility on demand by obfuscating passenger destination locations. Because there exists unused supply even at peak demand, multiple vehicles with noisy origin locations are assigned to passengers through an iterative Hungarian algorithm \cite{prorok_privacy-preserving_2017}.
In other work, authors provide a complementary method that provides robustness to noisy travel time estimates by making use of \emph{robot redundancy}~\cite{prorok_Redundant_2019a, prorok_Robust_2020, malencia_fair_2021}. In other words, the core idea of those works is to exploit redundancy to counter uncertainty and redeem performance. 
Although the idea of engineering robust systems with redundant resources is not new in a broad sense~\cite{kulturel-konak_Efficiently_2003a, ghare_Optimal_1969}, these works consider redundant mechanisms for the problem of mobile robot assignment under uncertainty, with arbitrary and potentially correlated probability distributions.

\subsubsection{Routing}
The quintessential routing problem is the Multi Traveling Salesperson Problem, where a team of agents must collectively visit a set of locations while minimizing the total amount of travel time \cite{lawler_traveling_1985, balasubramanian_risk-aware_2020}. Traditional approaches that assume a known fixed time of travel between any two cities, i.e., the graph edges have fixed costs, are fragile in scenarios that involve stochasticity, partial information, and modeling errors \cite{bektas_multiple_2006}. Pre-operative approaches to routing explicitly consider uncertainties such as robot failures, environmental dynamics, and changing task definitions when solving the Multiple Traveling Robot Problem \cite{sariel-talay_multiple_2009}.

A closely related routing problem is the Orienteering Problem, where the robot team seeks to maximize reward that can be collected at different nodes but are not required to visit all nodes (whereas traveling salespersons problems require all locations are covered) \cite{vansteenwegen_orienteering_2011}.
In the Multiple-Path Orienteering Problem, an adversary is capable of attacking a subset of the robot team which plans to maximize their reward under this threat \cite{shi_robust_2020}.
 In the Team Surviving Orienteers Problem, edge weights represent the probability of a robot surviving the traversal of that edge and there are constraints on each agents probability of survival over their full path \cite{jorgensen_team_2017}.

\subsubsection{Path-planning}
Recent works in resilient multi-agent path planning include planning under uncertain costs or times \cite{yakovlev_prioritized_2019, atzmon_robust_2020}, privacy \cite{zhang_privacy-preserving_2021}, and disruptions or attacks \cite{zhou_approximation_2019, chasparis_lp-based_2008}. In these pre-operative approaches, there is a known disturbance model and the plan is created with respect to this model such that the impact of the disturbance on the multi-robot system is minimized.  

Wagner and Choset \cite{wagner_path_2017} model agents with dilated sizes according to the uncertainty in their poses and then plan conflict free trajectories for these dilated agents. Whilst this work models uncertainty in the pose of the robots (which impact travel time), others directly model stochastic travel times by representing delays as either gamma distributions \cite{peltzer_stt-cbs_2020} or number of time steps \cite{atzmon_robust_2018}.

In these pre-operative approaches, incorporating uncertainty creates more expressive models than deterministic approaches, but there are still limits because these models are assumed to fully and correctly model disturbances. Additionally, pre-operative planning approaches create changes to the system even without the presence of a disruption. This more conservative approach readies the system for disruption and thus may perform suboptimally when disruptions do not occur.

\begin{openProblem}[Planning over uncertainty]
\label{op:plan_pre}
Pre-operative approaches to planning handle stochastic stressors through redundancy and risk-averse measures (e.g., conditional value at risk). While these methods provide robustness, and are complementary to each other in handling risk, they are conservative. 
More work is required to understand how best to plan under modeled uncertainty, e.g., recent works have only just begun studying risk adaptive approaches~\cite{rudolph_desperate_2021}.
\end{openProblem} 

\textbf{Intra-operative approaches.}
Multi-agent planning algorithms that adapt when a disturbance occurs are considered intra-operative approaches. Some intra-operative methods can identify disturbances and respond accordingly \cite{talebpour_adaptive_2019} while others identify degradation in system performance and adapt without knowledge of the source or type of disturbance \cite{ramachandran_resilience_2021, ramachandran_resilient_2020}. 

\subsubsection{Assignment}
Assignment algorithms that are able to adapt to real-time disturbances have received recent attention \cite{he_data-driven_2020, emam_adaptive_2020}. The high computational cost of calculating optimal assignments prevents the continuous calculation of assignments during runtime, therefore, intra-operative assignment methods often work to identify when re-assignment is worth the expense. Zhou et al.~\cite{zhou_risk-aware_2020} present an event-driven algorithm that recomputes an assignment only when certain conditions are met, e.g., there exists a path that is both shorter and has less uncertainty in travel time. Mayya et al.~\cite{mayya_resilient_2021} measure the degradation of robot capabilities due to disturbances such as fog or mud and re-assign robots to tasks that have experienced performance drops due to these degraded capabilities. 
In the language of our Section~\ref{sec:approaches} notation, this work measures capabilities $\stressors$ (which capture unmodeled disturbances) and re-assigns agent-task pairings $x$ to maximize the average task performance~$f$. 
\edited{Similarly, \cite{ramachandran_resilience_2019} reconfigure their heterogeneous team when failures occur to maintain a communication graph.}

In mobility on demand applications, it is expected that new demand is continuously added. To address the uncertainty in both future demand and vehicle supply, He et al.~\cite{he_data-driven_2020} present a receding horizon algorithm that solves a distributionally robust optimization problem at each time step to calculate vehicle load balancing while Alonso-Mora et al.\cite{alonso-mora_predictive_2017} decouple vehicle routing and passenger to account for uncertain future demand.

\subsubsection{Path-planning}
Similar to some pre-operative planning approaches, Zhou et al.~\cite{zhou_distributed_2020} assume the worst case adversarial attack is known to have at most $\alpha$ adversaries. However, instead of using fixed pre-planned routes, this intra-operative approach considers the worst-case attack at each time step when re-planning. Other intra-operative planning methods include navigation in human workspaces \cite{lo_towards_nodate} and warehouses \cite{honig_persistent_2019}. 

\begin{openProblem}[Off-task planning]
\label{op:off_task}
Many homogeneous and heterogeneous planning applications involve extent of redundancy, whether explicitly in the number of robots, or implicitly in robots with capabilities that are not in current use, or capability-complementarities that are not currently exploited. These redundant resources often lie dormant until needed. Rather, researchers should be asking how to account for these unused resources with respect to possible future use. For example, in mobility on demand, agents that are currently not assigned to riders can move to locations that decrease the uncertainty or wait times of possible future riders. Or in assignment, how do current coalitions impact the space of complementary resources available for possible future coalitions?
\end{openProblem}

\textbf{Post-operative approaches.}
Current work in multi-agent planning addresses planning for single instances/missions. However, future applications will involve robot teams completing repeated or continual missions (e.g., agricultural robotics or automated construction teams).
The resilience needed for the long-term success of these teams could be achieved through post-operative multi-agent planning. Rather than adapting the plan when encountering disturbance, as in an intra-operative approach, post-operative planning would adapt the algorithm over the long term based on the experience of repeated episodes of missions. For example, after servicing a single farm, a team could optimize $\stressors$, its model of environmental disturbances (e.g., mud), given the data from the plan $x$ that the team followed and the performance $f$ that it achieved.

A stream of pickup and delivery tasks creates a nearly endless multi-agent planning sequence. While pre- and intra-operative planning can adapt to disturbance during a given trip or day, these systems may be fragile over the long term as disturbance characteristics change. For example, robotic systems in agricultural settings must adapt to seasonal changes in disturbance as well as longer-term changes due to climate change. While post-operative planning methods are just beginning to receive attention (e.g., lifelong path planning~\cite{ma_lifelong_2019}), this nascent literature is far from creating the long-term autonomous robotic systems required to solve complex continuous missions across industries like transportation, manufacturing, and agriculture. To bridge this gap, new tools must be brought into multi-agent planning for such life-long learning, which has been identified by others as an open challenge in this space \cite{salzman_research_2020}.

\begin{openProblem}[Long-term survival]
\label{op:survival}
Current planning approaches focus on optimizing an objective function such as efficiency. However, for long duration applications such as oceanic and extraterrestrial exploration, survival is a more important objective than efficiency. While constraint based approaches consider single agent survival \cite{egerstedt_robot_2018}, researchers must study the `survival' of multi-agent teams, with questions such as: How do you define a multi-agent optimization objective for planning for survival? and how do individual agents' actions harm or benefit the survival of its teammates, and the survival of the team?
\end{openProblem}

\begin{openProblem}[Reliance on a world model]
\label{op:world_model}
Planning algorithms rely on a model of the world and its uncertainties. However, these approaches still fail because of model inaccuracies, such as black swan events. Researchers should investigate whether it is possible to create an accurate world model that captures black swan events and captures the ways the world model changes with respect to the robot system's actions. Or, if this model is not possible, how to be resilient to such unknowns despite having no model of them. If it is the latter, this planning problem is related to determining what it means for the robot system to be resilient (Open Problem 1).
\end{openProblem}



\vspace{-0.3cm}
\subsection{Control}
\label{subsec:control}
Multi-robot control strategies facilitate the organization of multiple robots to solve team-level, global tasks using local interaction rules. In this section, we review methods across control applications, including motion coordination, coverage control, formation control, and control for information gathering and surveillance. 
This section also includes the general problem of ensuring cooperative computation in multi-robot networks---a problem to which consensus-based approaches are often the answer~\cite{pasqualetti_Consensus_2012, cortes2017coordinated}. We focus on failure-prone or adversarial environments that lead to malfunctioning robots, or compromised communication channels, resulting in disruptions to the collective task. In other words, the stressors $\phi$ are typically misbehaving or adversarial robots, and protective (resilient) mechanisms are required to deal with (mis-)information being disseminated by these robots. 

\textbf{Pre-operative approaches.}
Cooperative control algorithms for robot teams are underpinned by the general assumption that all entities are indeed cooperative. 
This, however, cannot be generally guaranteed, as robots break, are compromised, or fail to process and interpret sensor information. As such, the robots themselves become the stressors ($\phi$) of the system. 
\subsubsection{Robot formations for resilient consensus} Building resilient formations ($x$) provides a precautionary means of overcoming such non-cooperative or faulty robots. This line of work borrows from seminal results in network science that define the notion of {resilient communication graphs} through a property widely referred to as \textit{r-robustness}~\cite{leblanc_Resilient_2013a,sundaram_Distributed_2011a}.
Pre-operative approaches apply these concepts to the domain of robotics by considering physically embedded multi-agent systems, with constrained communication and dynamic behaviors. 
The challenge is that testing these networks for \textit{r-robustness} is computationally demanding, and requires global knowledge of the topology.
By constructing \textit{resilient robot formations}, authors have demonstrated that distributed consensus algorithms converge safely, regardless of what non-cooperative robots are communicating~\cite{saldana_Triangular_2016a, saldana_Resilient_2017a, guerrero-bonilla_Formations_2017}. One of the earliest works in this domain demonstrates that the most basic resilient formation can be built via {triangular networks}~\cite{saldana_Triangular_2016a}. This topology has the attractive property that it can be constructed incrementally and verified in a decentralized manner, in polynomial time. Further work builds on this foundation:~\cite{guerrero-bonilla_Formations_2017} accounts for any number of non-cooperative robots,~\cite{guerrero-bonilla_Design_2018} presents sufficient conditions on the robot communication range to guarantee resilient consensus, and~\cite{guerrero-bonilla_Dense_2020} addresses three-dimensional space through cubic lattice-based formations.

\subsubsection{Pre-planned consensus policies} 
Mobile robot teams have communication graphs that, generally, vary over time, and as a consequence, the (rigid) resilient formations introduced in the prior paragraph are not necessarily maintained. 
By implementing connectivity management policies, one can ensure that resilience is guaranteed as the network topologies undergo change. To address this problem, authors use measures of the resilience of the communication graph, characterized by the algebraic connectivity~\cite{saulnier_Resilient_2017a, saldana_Resilient_2017a}, and by Tverberg partitions~\cite{park2017fault}. 
Resilience in the sense of \textit{r-robustness} has also been quantified probabilistically, as shown in~\cite{wehbe_Probabilistic_2021}, assuming that robot communication is subject to random failures that can be modeled using a probability distribution. Robots with access to such an estimate can evaluate how their future actions may affect the system's resilience.

When connectivity constraints cannot be satisfied due to hard physical constraints, we need to resort to additional methods. In~\cite{saldana_Resilient_2017a}, authors develop a {sliding window consensus protocol} that provably guarantees resilience when the union of communication graphs over a bounded period of time jointly satisfies robustness properties. Their policy selectively activates communication links to attain resilience while solving tasks that require the robot team to cover wide-spread areas (e.g., perimeter surveillance).
Other work considers the applications of formation control~\cite{usevitch2018resilient, guerrero2019realization} and leader-follower systems~\cite{usevitch2018finite}, whereby reference values are time-varying. Wang et al.~\cite{wang2019resilient} propose event-triggered update rules that can mitigate the influence of faulty or malicious agents.

We note that in aforementioned approaches, while the robot team is adaptive with respect to its communication topology and motion strategy, there is no adaptation with respective to the stressors $\phi$, i.e., the assumed number of non-cooperative or faulty robots is fixed---hence the \textit{pre-operative} classification of these approaches.

\subsubsection{Optimization-based trajectory control} 
The use of model-predictive (\eg receding horizon ) control for coordinating multi-robot systems consists of continuously finding paths for all robots in the system, such that a global objective is optimized (such as traffic throughput or overall fuel consumption), subject to certain constraints (\eg no vehicle's path collides with another path, nor with any fixed or moving obstacle). 
Coupled centralized approaches, which consider the joint configuration space of all involved vehicles, have the advantage of producing optimal and complete plans~\cite{kavraki:1996,kant:1986, schouwenaars:2001}. 
However, such methods rely on the fact that all vehicles cooperate in the globally determined plans~\cite{chen:2015, kant:1986}. Consequently, these approaches are notoriously brittle and susceptible to individual robot failures and non-cooperation. The work in~\cite{kuwata_Cooperative_2011a} shows that a monotonic cost reduction of global objectives can be achieved, even in non-cooperative settings. This feat, however, relies on the fact that neighboring vehicles reliably execute the agreed upon maneuvers up to a known error bound.

\subsubsection{Control barrier functions}
Reliability can also be achieved by defining a desirable subset of the robots' state space, and then generating control inputs that render this subset forward-invariant. Control barrier functions (CBFs)~\cite{borrmann_Control_2015} are a framework for establishing such forward invariance, hence providing the sought-after robustness. In one of the earliest works in this vein, Usevitch et al.~\cite{usevitch_Adversarial_2021} present a method for guaranteeing forward invariance of sets in sampled-data multi-agent systems in the presence of a set of worst-case adversarial agents (whereby the identities of the adversarial agents are known to the normal agents).
While CBFs provide a computationally efficient tool to guarantee safety in multi-agent environments, they generally assume perfect knowledge of other agents’ dynamics and behaviors (e.g.,~\cite{chen_Guaranteed_2020}).

\subsubsection{Combinatorial approaches}
Providing resilience to any number of robot drop-outs (e.g., due to denial-of-service attacks or failures) is a computationally challenging task, since one would need to account for all possible removals of robots from the joint planning task, which is a problem of combinatorial complexity. The work in~\cite{rabban2020improved} defines a resilient coverage maximization problem, in which the objective is to select a trajectory for each robot such that target coverage is maximized in the case of a worst-case failure of $\alpha$ robots. While it is assumed that at most $\alpha$ robots may fail, it is unknown which robots are going to fail. A similar assumption is made in~\cite{schlotfeldt2018resilient} for the case of active information gathering scenario, namely, multi-robot target tracking.

\subsubsection{Protective approaches} The topic of privacy remains poorly addressed within robotics at large. Yet, privacy can be an important facet of defence against active adversaries for many types of robotics applications. Using privacy as a defence mechanism is particularly relevant for collaborative robot teams, where individual robots assume different roles with varying degrees of specialization. As a consequence, specific robots may be critical to securing the system’s ability to operate without failure. The premise is that a robot’s motion may reveal sensitive information about its role within the team. 
Privacy preserving control methods, hence, tackle the problem of \textit{preventing an adversary from being able to distinguish the role} of one robot from that of another.
In~\cite{prorok_macroscopic_2016}, the authors consider collaboration across heterogeneous robot teams; their method  builds on the theory of differential privacy to quantify how easy it is for an adversary to identify the \textit{type} of any robot in the group, based on an observation of the robot group’s dynamic state. Note that, a similar, yet post-operative approach is taken in~\cite{zheng_adversarial_2020}.

\textbf{Intra-operative approaches.}
These approaches are dynamic, with decision variables $x$ adapting to changes perceived in $\phi$; e.g., measurements from non-attacked robots can be used to observe ongoing failures or newly perceived obstacles.

\subsubsection{Obstacle avoidance and adaptive navigation} In contrast to the coupled (centralized) trajectory control methods introduced above, {decentralized} approaches consider the generation of collision-free paths for individual robots that cooperate only with immediate neighbors~\cite{desaraju:2012,kuwata:2007}, or with no other vehicles at all~\cite{alonso:2012, vandenberg:2005, wang_Mobile_2020}. Hence, coordination is reduced to the problem of dynamically (and reciprocally) avoiding other vehicles (and obstacles), and can generally be solved without the use of explicit communication. Although such approaches are resilient to communication-based faults and attacks, the key disadvantage is that the optimality of global objectives (such as overall traffic efficiency) can generally not be guaranteed as robots follow ad-hoc policies.
The work in~\cite{csenbacslar2019robust} combines the best of both worlds, presenting a hybrid planning strategy employing both discrete planning and trajectory optimization with a dynamic receding horizon approach. Although pre-planned trajectories form the initial coordinated trajectory plan, their method allows for adaptation to dynamic changes, including newly appearing obstacles, robots breaking down, and imperfect motion execution. 
Also adapting to stressors in an online manner, the work in~\cite{cheng_Safe_2020} learns high-confidence bounds for dynamic uncertainties. This robust CBF formulation maintains safety with a high probability and adapts to the learned uncertainties.

\subsubsection{Security}
While most works addressing multi-robot fault tolerance through robust consensus policies make use of worst-case assumptions, approaches toward spoof detection make use of \textit{independent} physical channel observations (i.e., signal profiles), created by complex multi-path fading~\cite{wheeler2019switching,gil_guaranteeing_2017,renganathan_Spoof_2017}.
The methods differ, e.g.,~\cite{wheeler2019switching} determines which edges in the network to switch on or off over the evolution of the consensus in order to eliminate spoofed node influence, whereas~\cite{gil_guaranteeing_2017} assigns robot confidence values (signifying robot legitimacy).
The work in~\cite{mallmann-trenn_Crowd_2021} leverages a probabilistic measure of trustworthiness to find and eliminate adversarial robots in the presence of a Sybil attack.

\subsubsection{Combinatorial approaches}
The approaches introduced in the \textit{pre-operative} section above consider worst-case failures and over-provision for robustness (e.g., see~\cite{schlotfeldt2018resilient}). Differently, the work in~\cite{schlotfeldt2021resilient} continuously takes measurements from all non-attacked robots to observe ongoing failures (in an active information gathering task). The control algorithms, therefore, are calculated based on the actual observed stressors (i.e., attacked robots).
In a similar vein, Tzoumas et al.~\cite{tzoumas2018resilient} consider a similar scenario (i.e., fault-tolerant robot navigation with sensor scheduling), whereby at each time step, the algorithm selects system elements based on the history of inflicted attacks, deletions, or failures; this allows for guarantees of resiliency to any number of robot failures.

\begin{openProblem} [Design of signal complementarity for system resilience]
\label{op:signal_complementarity}
The commonality of many intra-operative approaches is that they leverage some independent signal, e.g., a physical observation or separate communication channel, which facilitates the online adaptation to stressors (e.g., see the need for an `eye-in-the-sky' in~\cite{saulnier_Resilient_2017a}). This, in turn, promotes the design of heterogeneous teams, that can provide the necessary complementary information. However, thus far, heterogeneous systems have been hand-designed, \edited{and their optimal control policies are hard to come by~\cite{prorok_Impact_2017a}. This compounds the problem of devising methods that incorporate heterogeneous modalities.}
\end{openProblem}

\begin{openProblem} [Co-design of control and communications]
\label{op:co_design_control_comms}
Time-varying and unreliable connectivity compounds the difficulty of 
\edited{resilient group coordination and control. Joint networking and control designs are needed that exploit evolving cognitive communications, provide self-healing network topology adaptation, and guarantee privacy and security. Enhanced perception-action-communication loop designs are also needed, that provide relevant signals and feedback to local agent controllers \cite{yang_grand_2018a, fink_Robust_2011}. }
\end{openProblem}

\textbf{Post-operative approaches.}
Learning-based methods have proven effective at designing robot control policies for an increasing number of multi-robot tasks, whereby Imitation Learning (IL) (e.g.,~\cite{tolstaya_Learning_2019a, li_Messageaware_2021}) and Reinforcement Learning (RL) (e.g.,~\cite{wang_Mobile_2020}) are currently the leading paradigms. In both cases, the learning procedure leverages information that is accumulated within robot neighborhoods, composing batches of data that are learnt from \textit{post-factum}.

\subsubsection{Multi-Agent Reinforcement Learning (MARL)}
Learning to interact in multi-agent systems is challenged by a non-stationarity of the environment, as agents learn concurrently to coordinate their actions, and continually change their decision-making policies~\cite{papoudakis_Dealing_2019}. An actor-critic method is presented in~\cite{lowe_MultiAgent_2017a} that successfully learns policies that require complex multi-agent coordination, discovering various physical and informational coordination strategies.
The work in \cite{omidshafiei_Deep_2017} introduces a decentralized single-task learning approach that is robust to concurrent interactions of teammates. It presents an approach for distilling single-task policies into a unified policy that performs well across multiple related tasks, without explicit provision of task identity.
The work in~\cite{zhang_Robust_2020a} studies the MARL problem with model uncertainty. The authors pose the problem as a robust Markov game, where the goal of all agents is to find policies such that no agent has the incentive to deviate, i.e., reach some equilibrium point, which is also robust to the possible uncertainty of the MARL model.
The work in~\cite{cheng_general_2021} proposes a framework that uses an epistemic logic to quantify trustworthiness of agents, and embed the use of quantitative trustworthiness values into control and coordination policies.
%

\subsubsection{Imitation Learning (IL)} The idea behind IL is to start with simple (small-scale) problems and use corresponding (optimal) solutions as examples to approach more complex, large-scale problems. This progression from example to application is crucial to mitigating the shortcomings of decentralized approaches in solving challenging multi-robot problems. 
Bridging the gap between the qualities of centralized and decentralized approaches, IL-based methods promise to find solutions that \textit{balance optimality and real-world efficiency}, as demonstrated in recent works, e.g.,~\cite{tolstaya_Learning_2019a, li_Graph_2020a, li_Messageaware_2021}. Although generalization to unseen cases has been successfully demonstrated, these approaches remain brittle due to their dependency on expert demonstrations during learning. 

\subsubsection{Graph Neural Networks (GNNs)} While centralized-training, decentralized-execution (CTDE)~\cite{oliehoek_Optimal_2008} is the \edited{typical} paradigm for multi-agent RL and multi-agent IL, the underlying machine learning framework can vary. 
\edited{Graph Neural Networks (GNNs) have} shown remarkable performance across a number of multi-robot problems~\cite{khan_2020, tolstaya_Learning_2019a, li_Graph_2020a, kortvelesy_ModGNN_2021,Talak21arxiv-neuralTree,Ravichandran21arxiv-RLwithSceneGraphs}. 
\edited{Graph nodes represent robots and edges model communication links between them~\cite{prorok_Graph_2018a}. GNNs provide a general learning framework for perception-action-communication loops that incorporates network topology, distributed processing, and control~\cite{hu_Scalable_2021}. Global state information can be distilled and shared through neighbor data exchange. 
GNNs, like conventional NNs, maybe susceptible to adversarial attack, e.g., malicious agents can learn to manipulate or outperform other agents sharing the same communication channel~\cite{blumenkamp_Emergence_2020b}. A countermeasure was proposed in~\cite{mitchell2020gaussian}, providing a probabilistic model that allows agents to compute confidence values quantifying the truthfulness of any given communication partner. This confidence can be used to suppress suspicious information. Yet, as noted in Open Problem~\ref{op:signal_complementarity}, this idea leans on information complementarity---future work should look to specifying the requirements needed, and guarantees that can be provided.
}

\subsubsection{Adversarial training}
Learning to deal with adversarial input or disruption during training is a promising approach to providing for resilience. In~\cite{zheng_adversarial_2020}, Zheng et al. leverage data-driven adversarial co-optimization, and design a mechanism that optimizes a flock's motion control parameters, such that the risk of flock \textit{leader identification} is minimized. This approach is reminiscent of the ideas in~\cite{prorok_macroscopic_2016} that aim to preserve role privacy. While the work in~\cite{blumenkamp_Emergence_2020b} first shows that an adversary can learn to exploit other agents' behaviors to better its own reward, it also shows that when the learning is alternated, cooperative agents are able to learn to re-coup their performance losses. This line of work was extended in~\cite{mitchell2020gaussian}, where a local filter is trained to detect implausible communication, allowing agents to cooperate more robustly.
The work in~\cite{stone_collaborative_1998} shows the necessity of performing both collaborative and adversarial learning, resulting in successful team performance that can withstand opponent attacks. 
Yet, recent work~\cite{lechner_Adversarial_2021} argues that adversarial training can introduce novel error profiles in robot learning schemes, and more work is required to fully understand how the method can be leveraged to for safety-critical applications.

\begin{openProblem} [Quick vs. slow learning]
\label{op:quick_v_slow}
\edited{Once deployed, a policy may become stale and should be updated based on newly collected data. If updated too soon, noisy data may lead to overfitting and poor generalization. } 
Conversely, not updating the policy often enough can lead to catastrophic failures and an inability to adapt.
\end{openProblem}

\begin{openProblem} [Unsupervised resilience learning]
\label{op:unsupervised_resilience}
Supervised learning, including reinforcement learning and imitation learning, requires apriori specification of rewards / cost functions, or access to expert data. 
Resilience, however, requires autonomous identification and diagnosis of failure to inform how robot policies and configurations should change post-operatively. It is currently unclear how this is to be achieved without supervisory intervention.
\end{openProblem}

\begin{openProblem} [Interpretability of multi-agent policies]
\label{op:interpret}
Time-varying and unpredictable connectivity patterns complexify the task of explaining and guaranteeing performance of multi-agent policies---literature on visualizing/interpreting multi-agent communication is sparse, with recent solutions designed specifically for the task at hand (e.g.,\cite{blumenkamp_Emergence_2020b}). 
\end{openProblem}

\definecolor{color_perception}{HTML}{da8d26}
\definecolor{color_planning}{HTML}{395eae}
\definecolor{color_control}{HTML}{b8364e}

\begin{figure*}[tb]
  \begin{subfigure}{0.3\textwidth}
    \centering
    \includegraphics[width = \textwidth]{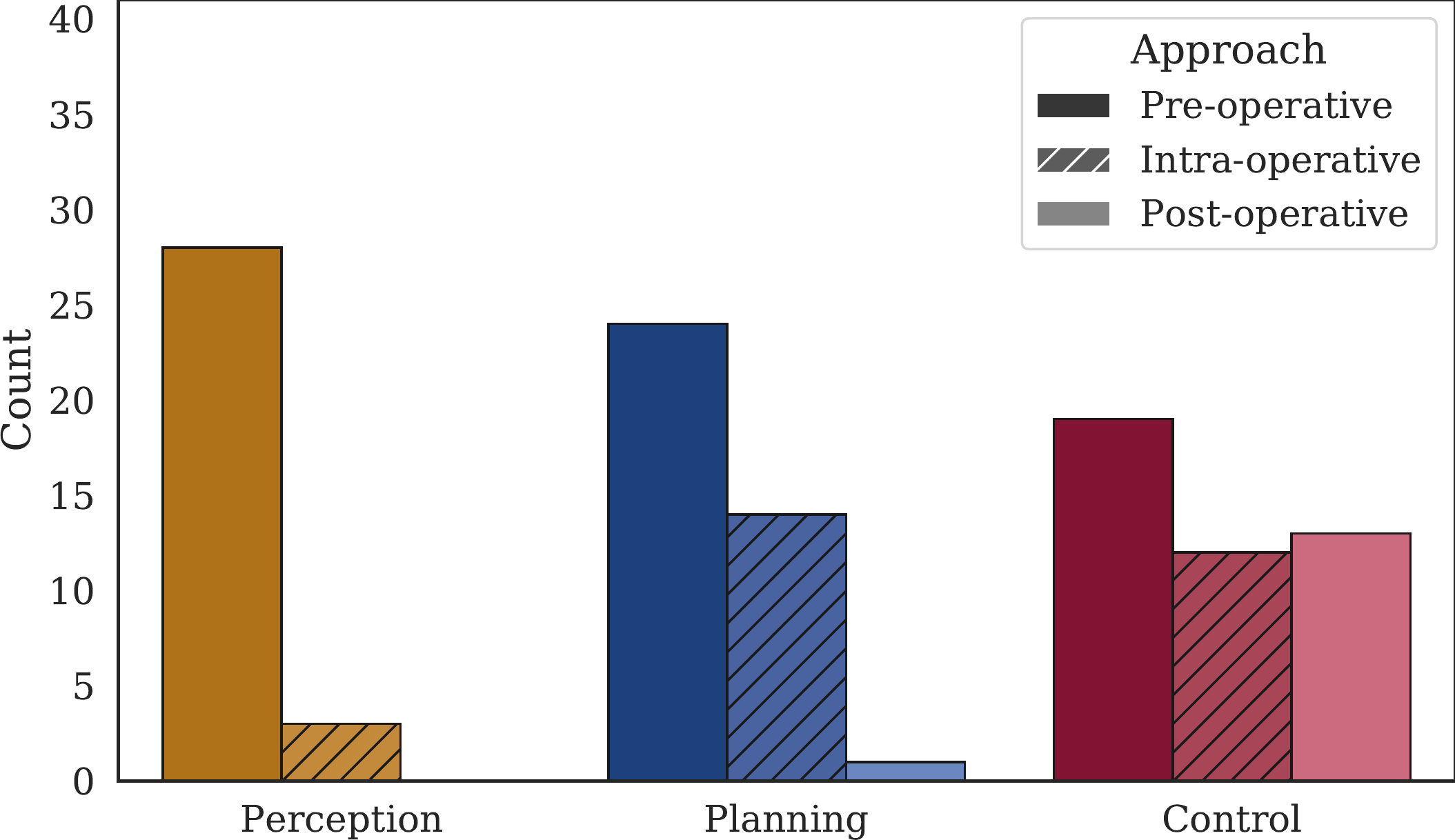}
    \caption{Approach} \label{fig:sum_approach}
  \end{subfigure}%
  \hspace*{\fill}   
  \begin{subfigure}{0.3\textwidth}
    \centering
    \includegraphics[width = \textwidth]{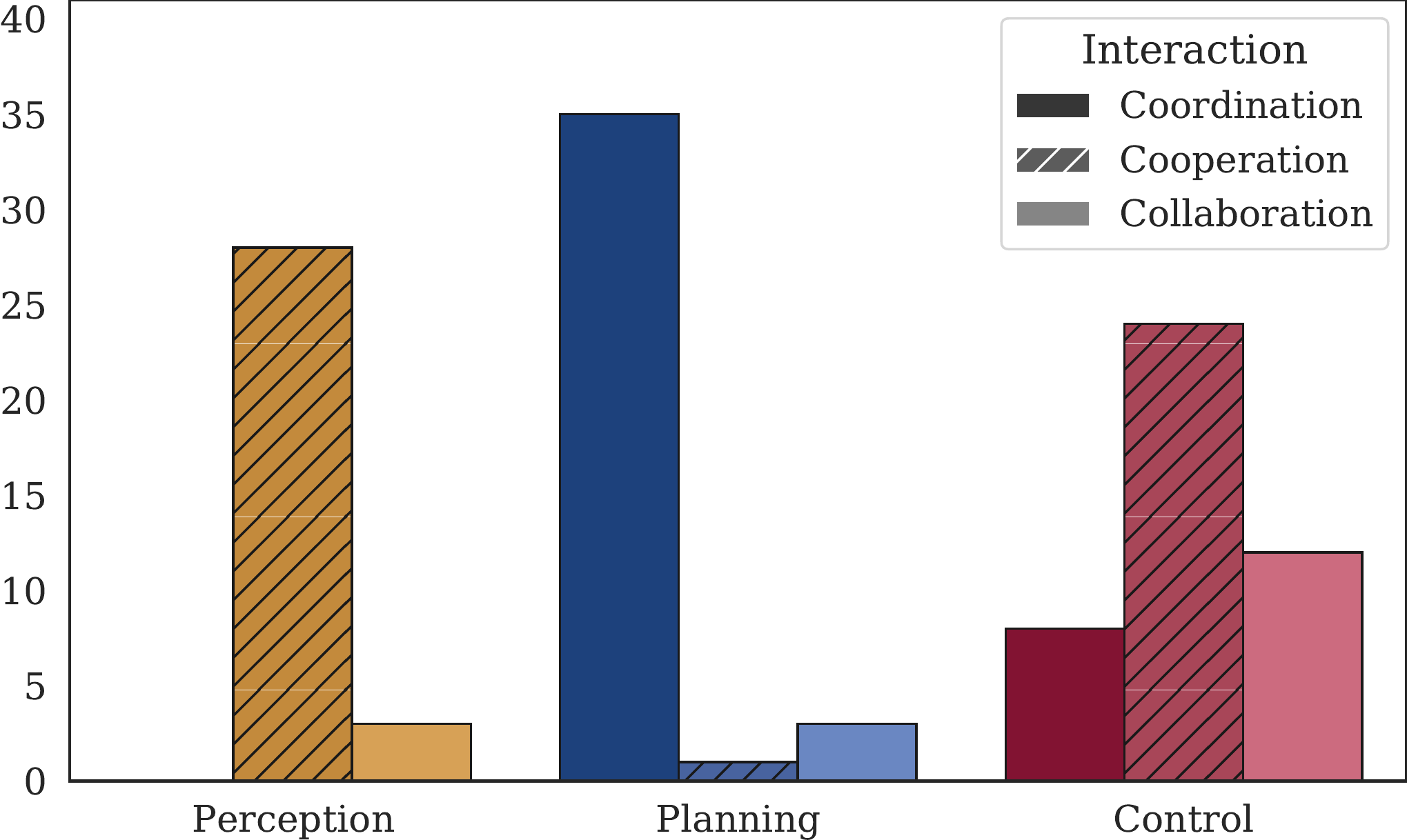}
    \caption{Interaction} \label{fig:sum_interact}
  \end{subfigure}%
  \hspace*{\fill}   
  \begin{subfigure}{0.3\textwidth}
    \centering
    \includegraphics[width = \textwidth]{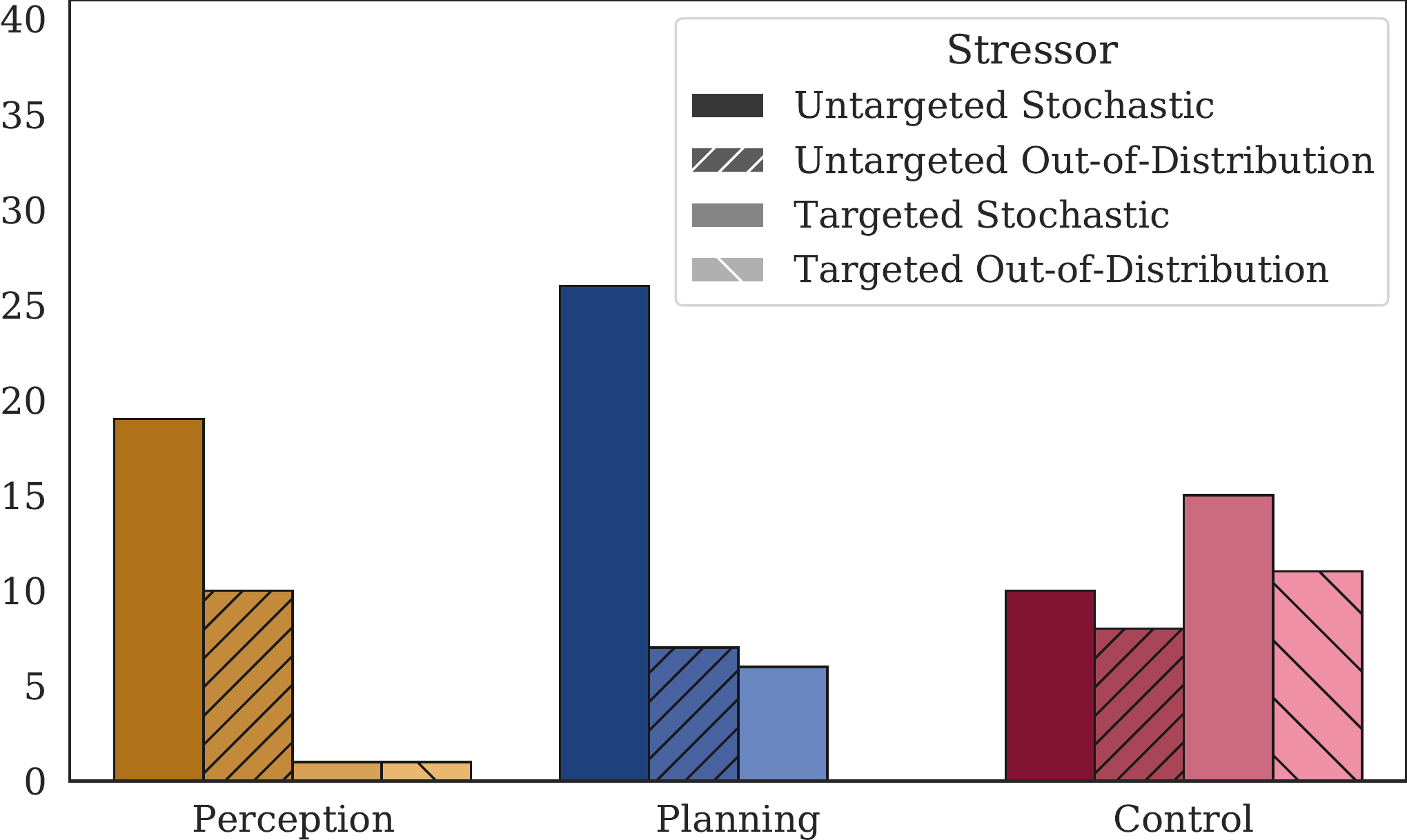}
    \caption{Stressors} \label{fig:sum_stressor}
  \end{subfigure}%
  \hspace*{\fill}   
    \caption{Summary statistics of Table \ref{taxonomy}. (\ref{fig:sum_approach}) the distribution of approaches for each domain, showing that research has focused on pre-operative approaches, especially so in perception. (\ref{fig:sum_interact}) the distribution of interaction type for each domain shows a dominance of coordination within planning research, cooperation in the perception domain, and a most evenly distributed for control. (\ref{fig:sum_stressor}) the summary statistics for stressors show that the controls domain has the most research on target stressors while perception and planning are dominated by untargeted and often stochastic stressors.
    }
    \label{fig:summary}
\end{figure*}

\renewcommand{\arraystretch}{1.5}

\begin{table*}[t]
\caption{Taxonomy of Resilience in Multi-Robot Systems, across domains: Perception (orange), Planning (blue), Control (red)}
\label{taxonomy}
\begin{tabular}{p{1.5cm} p{2.5cm} | p{6.5cm} | p{3cm} | p{2.5cm}}
\toprule
                     & Approach &                                                                                                                                                                                                                                                                                                                                                                                                                                                                                                                                                                                                                                                                                                                                                                                                                                                                                                                                                                                                                                                                                                                      Pre-operative &                                                                                                                                                                                                                                                                                    Intra-operative &                                                                                                                                                                                                                                                                       Post-operative \\
Interaction & Stressor &                                                                                                                                                                                                                                                                                                                                                                                                                                                                                                                                                                                                                                                                                                                                                                                                                                                                                                                                                                                                                                                                                                                                    &                                                                                                                                                                                                                                                                                                    &                                                                                                                                                                                                                                                                                      \\
\midrule
Coordination & Untargeted\newline Stochastic &  \textcolor{color_planning}{\cite{prorok_Robust_2020}},\textcolor{color_planning}{\cite{prorok_redundant_2019}},\textcolor{color_planning}{\cite{nam_analyzing_2017}},\textcolor{color_planning}{\cite{nam_when_2015}},\textcolor{color_planning}{\cite{jorgensen_team_2017}},\textcolor{color_planning}{\cite{peltzer_stt-cbs_2020}},\textcolor{color_planning}{\cite{wagner_path_2017}},\textcolor{color_planning}{\cite{atzmon_robust_2018}},\textcolor{color_planning}{\cite{prorok_privacy-preserving_2017}},\textcolor{color_planning}{\cite{atzmon_probabilistic_2020}},\textcolor{color_planning}{\cite{balasubramanian_risk-aware_2020}},\textcolor{color_planning}{\cite{malencia_fair_2021}},\textcolor{color_planning}{\cite{rudolph_desperate_2021}},\textcolor{color_planning}{\cite{ponda_distributed_2012}},\textcolor{color_planning}{\cite{shang_stochastic_2020}},\textcolor{color_planning}{\cite{zhang_privacy-preserving_2021}},\textcolor{color_planning}{\cite{yakovlev_prioritized_2019}},\textcolor{color_planning}{\cite{yang_algorithm_2020}},\textcolor{color_control}{\cite{chen_Guaranteed_2020}} &               \textcolor{color_planning}{\cite{atzmon_robust_2020}},\textcolor{color_planning}{\cite{he_data-driven_2020}},\textcolor{color_planning}{\cite{zhou_risk-aware_2020}},\textcolor{color_planning}{\cite{emam_adaptive_2020}},\textcolor{color_planning}{\cite{tai_prioritized_2019}} &                                                                                                                                                                         \textcolor{color_planning}{\cite{ma_lifelong_2019}},\textcolor{color_control}{\cite{li_Messageaware_2021}} \\
                     \rowcolor{gray!10}\cellcolor{white} &Untargeted\newline Out-of-Distribution &                                                                                                                                                                                                                                                                                                                                                                                                                                                                                                                                                                                                                                                                                                                                                                                                                                                                                                                                                                                                                                                                        \textcolor{color_planning}{\cite{prorok_Redundant_2019a}} &                                                    \textcolor{color_planning}{\cite{sariel-talay_multiple_2009}},\textcolor{color_planning}{\cite{mayya_resilient_2021}},\textcolor{color_planning}{\cite{honig_persistent_2019}},\textcolor{color_planning}{\cite{ramachandran_resilient_2020}} &                                                                                                                                                                                                                                                                                      \\
                     \arrayrulecolor{gray}\cline{2-5}\arrayrulecolor{black} & Targeted\newline Stochastic &                                                                                                                                                                                                                                                                                                                                                                                                                                                                                                                                                                                                                                                                                                                                                                                 \textcolor{color_planning}{\cite{shi_robust_2020}},\textcolor{color_planning}{\cite{liu_distributed_2021}},\textcolor{color_planning}{\cite{chasparis_lp-based_2008}},\textcolor{color_planning}{\cite{zhou_approximation_2019}},\textcolor{color_control}{\cite{schlotfeldt2018resilient}},\textcolor{color_control}{\cite{rabban2020improved}} &  \textcolor{color_planning}{\cite{zhou_distributed_2020}},\textcolor{color_planning}{\cite{talebpour_adaptive_2019}},\textcolor{color_control}{\cite{gil_guaranteeing_2017}},\textcolor{color_control}{\cite{schlotfeldt2021resilient}},\textcolor{color_control}{\cite{renganathan_Spoof_2017}} &                                                                                                                                                                                                                       \textcolor{color_control}{\cite{blumenkamp_Emergence_2020b}} \\
                     \rowcolor{gray!10}\cellcolor{white} &Targeted\newline Out-of-Distribution &                                                                                                                                                                                                                                                                                                                                                                                                                                                                                                                                                                                                                                                                                                                                                                                                                                                                                                                                                                                                                                                                                                                                    &                                                                                                                                                                                                                                                                                                    &                                                                                                                                                                                                                                                                                      \\
\hline Cooperation & Untargeted\newline Stochastic &                                                                                        \textcolor{color_control}{\cite{kuwata_Cooperative_2011a}},\textcolor{color_control}{\cite{wehbe_Probabilistic_2021}},\textcolor{color_perception}{\cite{Tian18rss}},\textcolor{color_perception}{\cite{Giamou18icra}},\textcolor{color_perception}{\cite{Barooah07csm}},\textcolor{color_perception}{\cite{Schenato07ieee}},\textcolor{color_perception}{\cite{Roumeliotis02tra}},\textcolor{color_perception}{\cite{Thrun03isrr}},\textcolor{color_perception}{\cite{Howard06ieee}},\textcolor{color_perception}{\cite{Carlone10jirs-multiRobotSLAM}},\textcolor{color_perception}{\cite{Andersson08icra}},\textcolor{color_perception}{\cite{Kim10icra}},\textcolor{color_perception}{\cite{Bailey11icra}},\textcolor{color_perception}{\cite{Lazaro11icra}},\textcolor{color_perception}{\cite{Aragues11icra-distributedLocalization}},\textcolor{color_perception}{\cite{Cunningham13icra}},\textcolor{color_perception}{\cite{Choudhary17ijrr-distributedPGO3D}},\textcolor{color_perception}{\cite{Tian19arxiv-distributedSEsync}} &                                                                                                                                          \textcolor{color_control}{\cite{csenbacslar2019robust}},\textcolor{color_control}{\cite{kuwata:2007}},\textcolor{color_control}{\cite{cheng_Safe_2020}} &                                                                                                                                                                                                                                                                                      \\
                     \rowcolor{gray!10}\cellcolor{white} &Untargeted\newline Out-of-Distribution &                                                                                                                                                                                                                                                                                                                                                                                                                                                          \textcolor{color_control}{\cite{usevitch2018resilient}},\textcolor{color_perception}{\cite{Tardioli15iros}},\textcolor{color_perception}{\cite{Cieslewski18icra}},\textcolor{color_perception}{\cite{Cieslewski17ral-bow}},\textcolor{color_perception}{\cite{Cieslewski17mrs-netvlad}},\textcolor{color_perception}{\cite{Lajoie20ral-doorSLAM}},\textcolor{color_perception}{\cite{Liu20cvpr-when2com}},\textcolor{color_perception}{\cite{Wu19iccv-RLVideoRecognition}},\textcolor{color_perception}{\cite{Mousavi19iros-multiAgentImageClassification}},\textcolor{color_perception}{\cite{Dong15icra}},\textcolor{color_perception}{\cite{tian_Asynchronous_2020}} &                                                                                                                                                                                    \textcolor{color_planning}{\cite{ramachandran_resilience_2021}},\textcolor{color_control}{\cite{alonso:2012}} &                                                                                                                                                                                                                             \textcolor{color_control}{\cite{bossens_Rapidly_2021}} \\
                     \arrayrulecolor{gray}\cline{2-5}\arrayrulecolor{black} & Targeted\newline Stochastic &                                                                                                                                                                                                                                                                                                                                                                                                                                                                                                                                                                                                                                                                                                                                                                                                                                                                                                        \textcolor{color_control}{\cite{leblanc_Resilient_2013a}},\textcolor{color_control}{\cite{wang2019resilient}},\textcolor{color_control}{\cite{usevitch_Adversarial_2021}},\textcolor{color_control}{\cite{park2017fault}} &                                                             \textcolor{color_perception}{\cite{lajoie_DOORSLAM_2020}},\textcolor{color_control}{\cite{sundaram_Distributed_2011a}},\textcolor{color_control}{\cite{wheeler2019switching}},\textcolor{color_control}{\cite{tzoumas2018resilient}} &                                                                                                                                                                                                                            \textcolor{color_control}{\cite{lowe_MultiAgent_2017a}} \\
                     \rowcolor{gray!10}\cellcolor{white} &Targeted\newline Out-of-Distribution &                                                                                                                                                                                                                                                                                                                                                                                                                                                                                                                                                                                                                                                                                                                                                                \textcolor{color_control}{\cite{saulnier_Resilient_2017a}},\textcolor{color_control}{\cite{saldana_Resilient_2017a}},\textcolor{color_control}{\cite{saldana_Triangular_2016a}},\textcolor{color_control}{\cite{guerrero-bonilla_Dense_2020}},\textcolor{color_control}{\cite{guerrero-bonilla_Design_2018}},\textcolor{color_control}{\cite{usevitch2018finite}} &                                                                                                                                                                                                                                                \textcolor{color_perception}{\cite{Tzoumas18cdc}} &                                                                                                                                                                    \textcolor{color_control}{\cite{mitchell2020gaussian}},\textcolor{color_control}{\cite{zheng_adversarial_2020}} \\
\hline Collaboration & Untargeted\newline Stochastic &                                                                                                                                                                                                                                                                                                                                                                                                                                                                                                                                                                                                                                                                                                                                                                                                                 \textcolor{color_planning}{\cite{sharma_risk-aware_2020}},\textcolor{color_control}{\cite{prorok_Impact_2017a}},\textcolor{color_control}{\cite{carlone_Robot_2019b}},\textcolor{color_perception}{\cite{Forster13iros-airGroundLocalization}},\textcolor{color_perception}{\cite{Michael14fr-airGroundMapping}} &                                                                                                              \textcolor{color_perception}{\cite{alonso-mora_predictive_2017}},\textcolor{color_planning}{\cite{lo_towards_nodate}},\textcolor{color_control}{\cite{ramachandran_resilient_2020}} &                                                                                                                                                                                                                                                                                      \\
                     \rowcolor{gray!10}\cellcolor{white} &Untargeted\newline Out-of-Distribution &                                                                                                                                                                                                                                                                                                                                                                                                                                                                                                                                                                                                                                                                                                                                                                                                                                                                                                                                                                                                                                                                                                                                    &                                                                                                                                                                                                                                  \textcolor{color_planning}{\cite{ramachandran_resilience_2019}} &  \textcolor{color_control}{\cite{dasilva_Agents_2020a}},\textcolor{color_control}{\cite{papoudakis_Dealing_2019}},\textcolor{color_control}{\cite{zhang_Recent_2021}},\textcolor{color_control}{\cite{zhang_Robust_2020a}},\textcolor{color_control}{\cite{omidshafiei_Deep_2017}} \\
                     \arrayrulecolor{gray}\cline{2-5}\arrayrulecolor{black} & Targeted\newline Stochastic &                                                                                                                                                                                                                                                                                                                                                                                                                                                                                                                                                                                                                                                                                                                                                                                                                                                                                                                                                                                                                                                                                                                                    &                                                                                                                                                                                                                                                                                                    &                                                                                                                                                                                                                         \textcolor{color_control}{\cite{stone_collaborative_1998}} \\
                     \rowcolor{gray!10}\cellcolor{white} &Targeted\newline Out-of-Distribution &                                                                                                                                                                                                                                                                                                                                                                                                                                                                                                                                                                                                                                                                                                                                                                                                                                                                                                                                                                                                                                                                        \textcolor{color_control}{\cite{prorok_macroscopic_2016}} &                                                                                                                                                                                                                                      \textcolor{color_control}{\cite{mallmann-trenn_Crowd_2021}} &                                                                                                                                                                                                                               \textcolor{color_control}{\cite{cheng_general_2021}} \\
\bottomrule
\end{tabular}
\end{table*}



\newcommand{\tick}{\ding{51}}
\begin{table*}[t]
\caption{Classification of Open Problems \label{tab:open_problems}}
\begin{tabular}{p{1cm} | p{1cm} p{1cm} p{1cm} | p{1.5cm} p{1.5cm} p{1.5cm} | p{1cm} p{1.5cm} p{1cm} p{1.5cm}}
\toprule
& \multicolumn{3}{c}{Approach} & \multicolumn{3}{c}{Interaction} & \multicolumn{2}{c}{Untargeted} & \multicolumn{2}{c}{Targeted} \\
Problem & Pre-operative & Intra-operative & Post-operative & Coordination & Cooperation & Collabation & Stochastic & Out-of-Distribution & Stochastic & Out-of-Distribution \\ \hline
OP~\ref{op:resilience_measure} & \tick & \tick & \tick & \tick & \tick & \tick & & \tick & \tick & \tick  \\ 
\hline
\rowcolor{gray!10!}OP~\ref{op:team_learning} & \tick &  &  &  & \tick & \tick &  & \tick &  & \tick \\
OP~\ref{op:scene_understanding} & \tick &  &  &  & \tick & \tick &  & \tick &  & \\
\rowcolor{gray!10!}OP~\ref{op:failure_reasoning} &  & \tick &  &  & \tick &  &  & \tick &  & \\
OP~\ref{op:active_perception} &  & \tick &  &  & \tick & \tick &  & \tick &  & \\
\rowcolor{gray!10!}OP~\ref{op:tuning_reconfig} &  &  & \tick &  & \tick &  &  & \tick &  & \tick \\ 
\hline
OP~\ref{op:plan_pre} & \tick &  &  & \tick & \tick &  & \tick &  & \tick & \\
\rowcolor{gray!10!}OP~\ref{op:off_task} &  & \tick &  & \tick & \tick & \tick & \tick &  & \tick & \\
OP~\ref{op:survival} &  & \tick & \tick & \tick & \tick &  &  & \tick &  & \tick \\
\rowcolor{gray!10!}OP~\ref{op:world_model} &  & \tick & \tick & \tick & \tick & \tick &  & \tick & & \tick \\ 
\hline
OP~\ref{op:signal_complementarity} &  & \tick &  &  &  & \tick &  & \tick &  & \tick \\
\rowcolor{gray!10!}OP~\ref{op:co_design_control_comms} &  & \tick &  &  & \tick & \tick &  & \tick & & \tick \\
OP~\ref{op:quick_v_slow} &  &  & \tick & \tick & \tick & \tick & \tick &  & \tick &  \\
\rowcolor{gray!10!}OP~\ref{op:unsupervised_resilience} &  &  & \tick & \tick & \tick & \tick & \tick & \tick & \tick & \tick \\
OP~\ref{op:interpret} &  &  & \tick & &  & \tick &  & \tick &  & \tick \\
\hline
\rowcolor{gray!10!}OP~\ref{op:co-design} & \tick & & \tick & \tick & \tick &  & \tick & \tick  &  & \\
OP~\ref{op:co-opt} & \tick & & \tick & \tick & \tick &  & \tick & \tick &  & \\
\bottomrule
\end{tabular}
\end{table*}



\vspace{-0.3cm}
\subsection{Other Applications}
\label{subsec:others}

\subsubsection{Robot Co-design} 
Co-design problems aim at jointly designing sensing, computation, control and other algorithmic aspects 
that enable robots to perform a given task. Most of the approaches in this section are pre-operative, in the sense that they design for the worst case, but some evolutionary approaches can be considered post-operative since they evolve the system design after multiple executions.

Traditional control-theoretic approaches study sensor selection~\cite{Faming11icmsi,Joshi09tsp-sensorSelection,Gupta06automatica,Leny11tac-scheduling,Jawaid15automatica-scheduling,Zhao16cdc-scheduling,Tzoumas16acc-sensorScheduling,Carlone18tro-attentionVIN,Summers16tcns-sensorScheduling,Nozari17acc-scheduling,Summers17arxiv,Summers17arxiv2,Golovin10icipsn-sensorSelection}, while more modern techniques co-design sensing and control~\cite{tanaka15cdc-sdplqg,Tatikonda04tac-limitedCommControl,Tzoumas18acc-sLQG,Tzoumas20tac-sLQG}. 
A main limitation of this line of work is that the pursuit of theoretical guarantees limits these 
papers to focus on linear dynamical systems, a representation that struggles to capture the nonlinear and 
possibly discrete nature of perception and sensing in real-world robotics.
Evolutionary approaches~\cite{Lipson00nature-codesign,Hornby03tro-codesign,Lipson16ecal-codesign,Cheney18jrsi-codesign}
 provide a powerful paradigm that can indeed be understood in terms of Reinforcement Learning; 
 this approach has not been applied to sensing and perception aspects, due to the size of the search space (\eg choice of algorithms, parameters, and computation) and the difficulty of designing differentiable perception modules (\eg~\cite{Brachmann17CVPR-DSAC}).
 Similar considerations hold for modular languages and modularity-based approaches~\cite{Mehta15jmr-codesign,Hornby03tro-codesign,Ramos18icae}, where perception is typically simplified to reduce the size of the library or language and make the design tractable.
 Only a few optimization-based co-design approaches have explicitly tackled sensing and perception.
 Among those, Zhang~\etal~\cite{Zhang17rss-vioChip} investigate hardware-and-algorithms co-design 
 for visual-inertial odometry, and provide a heuristic approach to explore the search space.
 Zardini~\etal~\cite{Zardini21arxiv-codesign} leverage Censi's monotone co-design theory~\cite{Censi15arxiv-codesign} to design  hardware and software for an autonomous vehicle. 
 The work~\cite{Carlone19icra-codesign} designs sensing and hardware for a multi-robot team in charge of a 
 collective transport task using integer linear programming. 

\begin{openProblem}[Multi-Robot Co-design]
The literature on co-design is in its infancy, and the current tools for automated design 
still fall short from providing a satisfactory design tool for real-world robotics problems. 
E.g., none of the existing approaches is able to tame the complexity of a modern SLAM 
pipeline, due to their scalability limitations and underlying assumptions. 
In particular, co-design approaches neglect resilience altogether (while robustness is investigated in~\cite{Censi17ral-codesign}) and only a few 
involve multi-robot systems~\cite{Carlone19icra-codesign}.
\label{op:co-design}
\end{openProblem}

\subsubsection{Co-optimization of environment and multi-robot policies}
Current approaches to the design of mobile robot systems consider the environment as a fixed constraint~\cite{gombolay_Fast_2013a, smith_Estimating_1990a, prorok_Multilevel_2011a}. In the case of navigation, structures and obstacles must be circumnavigated; in this process, mobile agents engage in negotiations for right-of-way, driven by local incentives to minimize individual delays. Even in cooperative systems, environmental constraints can lead to dead-locks, live-locks, and prioritization conflicts~\cite{bennewitz_Finding_2002a, jager_Decentralized_2001a}. 
Despite the obvious influence of spatial constraints on agent interactions~\cite{boudet_collections_2021b}, the optimization of mobile robot systems and their immediate environment has, traditionally, been \textit{disjoint}, and little thought is given to what would make an artificial environment \textit{conducive} to effective and efficient collaboration, cooperation and coordination within mobile robot systems. 

As we progress with automated, roboticized systems, we must jointly re-evaluate the shape, form, and function of the environments that we operate in. Ultimately, this approach will allow us to overcome incremental research results, based on solutions that consider the environment as a fixed constraint, to provide for robustness and resilience in a holistic way. 

\begin{openProblem}[Co-optimization of robots and their environment]
Concurrent optimization of robot policies and environment they operate in has received little attention thus far (although evidence suggests significant benefits, e.g.,~\cite{cap_Asynchronous_2013,saunders_Teaching_2006a}). Such approaches are particularly applicable in man-made workspaces (e.g., factories, warehouses and urban settings), especially when stressors originate in the environment.
\label{op:co-opt}
\end{openProblem}


\vspace{-0.4cm}
\section{Discussion} 
\label{sec:open}



\edited{
Figure~\ref{fig:summary} summarizes the papers surveyed in Table~\ref{taxonomy}. Fig.~\ref{fig:sum_approach} shows that while the area of Control is well balanced across pre-, intra-, and post-operative approaches, Perception and Planning are skewed towards the pre-operative. 
Fig.~\ref{fig:sum_interact} shows how each domain has a prevailing mode of interaction---cooperation in Perception, coordination in Planning, and cooperation in Control (although, again, Control is the most balanced of the three domains). 
And lastly, Fig.~\ref{fig:sum_stressor} shows the distribution of work done across stressor types. Unsurprisingly, untargeted stochastic stressors are prevalent in Perception and Planning. Control features contributions across all stressor types; this can be traced back to the origins of research on Byzantine and non-cooperative networks, which are closely intertwined with the Controls community.

Our overview of open problems, see Table~\ref{tab:open_problems}, indicates that cooperation and collaboration are `harder' than coordination, which is understandable, as cooperation and collaboration tend to involve more problem dimensions and can be of complex combinatorial nature;
it also shows that of all stressors, out-of-distribution stressors need more attention, both untargeted as well as targeted.
Finally, fewer pre-operative methods are considered as part of open problems than intra- and post-operative approaches; this is reminiscent of recent focus less on a priori designs (pre-operative), rather more on designs that are adaptive (intra-operative) and learnable (post-operative).
}

\subsection{Grand Challenges}
\label{subsec:open_perception}

\textit{Introspective,~Resilient,~Multi-Robot~High-level~Understanding:} 
We believe a grand challenge in multi-robot perception is to develop multi-robot teams that can build a human-level shared representation of the environment (encompassing geometry, semantics, relations among entities in the scene, and more) in real-time and under computation and communication constraints. A second grand challenge is the design of truly resilient perception algorithms: we believe that the first step towards this goal is to develop \emph{introspection techniques} that can reason over failures, rather than just trying to avoid failures at all costs; the second step would then be to understand how automated system tuning and reconfiguration would impact the system performance in response to a failure.


\textit{Redundancy vs Complementarity:}
\edited{The open problems in Sections~\ref{subsec:planning} and~\ref{subsec:control} highlight the challenge of including adequate levels of complementarity and/or redundancy in system designs, for example through the provision of orthogonal sensing capabilities, distributed across the robot team, or redundant numbers of robots.
This pre-operative approach relates to the idea of \textit{anticipatory} resilience (cf. Fig.~\ref{fig:resilience_vs_robustness}). While redundancy is reminiscent of over-provisioning (and classical notions of robustness), its purpose in this context is to target unexpected disruptions (in contrast to modeled disruptions).
The grand challenge consists of devising foundational methods that inform which capabilities are to be integrated, and through which interaction paradigms.
}

\textit{Inter-disciplinary Resilience:}
\edited{The works in Table~\ref{taxonomy} highlight resilience research in the domains of perception, planning, and control. However, the complexity and inter-disciplinary nature of future applications of multi-agent systems requires that we investigate resilience at the \textit{intersection} of robotics domains.
A grand challenge is to investigate the complex interplay and second order effects of stressors across the domains of perception, planning, and control. For example, failures due to stressors on an agent's perception could be addressed through planning by leveraging a heterogeneous teammate's complementary sensor system that is more resilient to the targeted stressor that is encountered (see Open Problem~\ref{op:signal_complementarity}).
A second grand challenge, aligned with Open Problem \ref{op:resilience_measure}, is to develop interdisciplinary measures of resilience.}
\vspace{-0.4cm}
\subsection{Survivorship Bias}
\label{subsec:bias}

Current research practice and publication standards pressure the community to report successes only, which leads to a culture wherein failures and mistakes may be poorly documented, undisclosed, and consequently, not discussed publicly. Operating in this manner reinforces a \textit{survivorship bias}\footnote{\url{https://en.wikipedia.org/wiki/Survivorship_bias}}, which stymies learning from errors and controversially, leads to fragile designs~\cite{mangel_Abraham_1984}. Changes in publication culture, including more venues targeting negative results, would help accelerate progress towards resilient solutions.







\vspace{-0.3cm}
\section{Acknowledgments} \label{sec:acks}
We thank James Paulos for fruitful discussions in the early stages of this work.






\vspace{-0.3cm}
\bibliographystyle{bibfiles/IEEEtran}
\bibliography{bibfiles/refs_ap,bibfiles/refs_lc,bibfiles/refs_mm}


\end{document}